\documentclass[journal]{IEEEtran}
\usepackage{amsmath,amsfonts}
\usepackage{algorithmic}
\usepackage{algorithmic}
\usepackage{algorithm}
\usepackage{array}
\usepackage[caption=false,font=footnotesize,labelfont=rm,textfont=rm]{subfig}
\usepackage{textcomp}
\usepackage{stfloats}
\usepackage{url}
\usepackage{verbatim}
\usepackage{graphicx}
\usepackage{cite}
\usepackage{multirow}
\usepackage{booktabs}
\usepackage{tabularx}
\usepackage{amsfonts}

\newcommand{\etal}{{\em et al\,. }}       
\newcommand{\eg}{{\em e.g., }}           
\newcommand{\ie}{{\em i.e., }}           
\usepackage{makecell}

\usepackage{xcolor}

\usepackage{xcolor}
\definecolor{lightgray}{gray}{0.6}

\begin{document}
\title{Multi-level Reliable Guidance for Unpaired Multi-view Clustering}
\author{Like~Xin, Wanqi~Yang*, Lei~Wang, Ming~Yang
\thanks{This work is supported by the National Natural Science Foundation of China (Nos. 62476136, 62276138), the Qing Lan Project of Jiangsu Province, China, and the Postgraduate Research \& Practice Innovation Program of Jiangsu Province (KYCX24\_1865). Corresponding author: Wanqi Yang.}
\thanks{Like Xin is with the School of Mathematical Sciences, Nanjing Normal University, Nanjing, 210046, China.  \protect (e-mail: xinlike94@gmail.com).}
\thanks{Wanqi Yang and Ming Yang are with the School of Computer and Electronic Information, Nanjing Normal University, Nanjing, 210046, China. \protect (e-mail: yangwq@njnu.edu.cn, myang@njnu.edu.cn).}
\thanks{Lei Wang is with the School of Computing and Information Technology, University of Wollongong, Australia. \protect (e-mail: leiw@uow.edu.au).}
}
\markboth{IEEE Transactions on Neural Networks and Learning Systems}%
{Shell \MakeLowercase{\textit{et al.}}: Multi-level Reliable Guidance for Unpaired Multi-view Clustering}

\maketitle

\begin{abstract}
In this thesis, we address the challenging problem of unpaired multi-view clustering (UMC), which aims to achieve effective joint clustering using unpaired samples observed across multiple views. Traditional incomplete multi-view clustering (IMC) methods typically rely on paired samples to capture complementary information between views. However, such strategies become impractical in the UMC due to the absence of paired samples. Although some researchers have attempted to address this issue by preserving consistent cluster structures across views, effectively mining such consistency remains challenging when the cluster structures {with low confidence}.
Therefore, we propose a novel method, Multi-level Reliable Guidance for UMC (MRG-UMC), which integrates multi-level clustering and reliable view guidance to learn consistent and confident cluster structures from three perspectives.
Specifically, inner-view multi-level clustering exploits high-confidence sample pairs across different levels to reduce the impact of boundary samples, resulting in more confident cluster structures. Synthesized-view alignment leverages a synthesized-view to mitigate cross-view discrepancies and promote consistency. Cross-view guidance employs a reliable view guidance strategy to enhance the clustering confidence of poorly clustered views.
These three modules are jointly optimized across multiple levels to achieve consistent and confident cluster structures.
Furthermore, theoretical analyses verify the effectiveness of MRG-UMC in enhancing clustering confidence.
Extensive experimental results show that MRG-UMC outperforms state-of-the-art UMC methods, achieving an average NMI improvement of 12.95\% on multi-view datasets. {The source code is available at: https://anonymous.4open.science/r/MRG-UMC-5E20.}
\end{abstract}


\section{Introduction}
\IEEEPARstart {D}{ata} often exhibit diverse characteristics and representations, referred to as \textbf{multi-view data} \cite{xu2013survey, yang2015mrm, yang2022exploiting, yang2021corporate}. 
In the real world, some instances lack data in one or more views, resulting in \textbf{incomplete multi-view data} \cite{2015Multi, yang2018semi}. For example, in an image dataset, certain images may be missing specific visual or textual features.
A more extreme yet realistic scenario arises when no paired observed samples exist between views \cite{10149819}, rendering the multi-view data as \textbf{unpaired multi-view data}. 
For example, in multi-camera surveillance systems, individual cameras may operate intermittently due to energy-saving or maintenance, resulting in unpaired multi-view data \cite{10149819,scl-UMC, RGUMCxlk}. Additionally, in a video, if we view a visible video and accompanying narrator text as separate views, the description may become disjointed from the visual content due to delays, resulting in the occurrence of unpaired multi-view data \cite{wen2023unpaired}.

{Clustering \cite{yang2018multi} serves as a potent technique for uncovering inherent patterns within data, segmenting samples into distinct clusters autonomously without supervised information. 
For the three types of datasets mentioned above, three downstream clustering tasks—multi-view clustering (\textbf{MC} \cite{fang2023comprehensive}), incomplete multi-view clustering (\textbf{IMC} \cite{lin2021completer}), and unpaired multi-view clustering (\textbf{UMC} \cite{scl-UMC})—naturally emerge.} 
Specifically, UMC is more challenging than MC and IMC due to the absence of paired samples. We summarize the differences between UMC and these general clustering methods (MC and IMC) from the following two perspectives. 
i) High practicality. Complete multi-view clustering assumes that all samples are observed across views, and incomplete multi-view clustering allows for partially observed samples. However, such ideal conditions are rarely met in real-world applications.
For instance, in autonomous driving, LiDAR can detect distant vehicles, but cameras may miss them due to range limitations, making object association across views challenging.
On different social platforms (\eg X and LinkedIn), user data are not shared due to privacy concerns and business constraints, hindering personalized recommendations on emerging platforms.
In finance, payment records from platforms such as Alipay, PayPal, and Apple Pay are mutually exclusive, as each transaction occurs only once on a specific platform, complicating fraud detection.
Given these situations, UMC offers a more practical solution.
{ii) Independence from paired samples. Most complete and incomplete multi-view clustering methods rely on fully or partially paired samples to establish correspondences across views. In the absence of such correspondences, they often degenerate into independent single-view clustering, losing the ability to exploit cross-view complementarity. In contrast, our method assumes a consistent cluster structure across views, enabling the modeling of cross-view relationships without requiring paired samples, thereby offering greater flexibility and effectiveness for UMC.}

Recently, some methods have leveraged consistent cluster structures to connect different views in UMC. For example, Yang \etal \cite{10149819} leverage the covariance matrix to align different views. Xin \etal \cite{scl-UMC} use inner-view and cross-view contrastive learning to learn the consistent cluster structure. In addition, Xin \etal \cite{RGUMCxlk} leverage the view with the best clustering performance to align the other views. 
However, in UMC, learning consistent cluster structures across views often overlooks issues such as noise, boundary samples and ambiguous clustering initialization, which may lead to cluster structures with low confident.

\begin{figure}[!t]
\centering
\includegraphics[width=0.98\columnwidth]{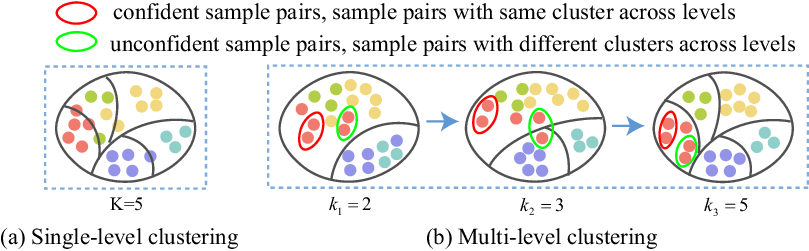}
\caption{The difference between single-level and multi-level clustering. In single-level clustering, some boundary samples typically emerge when there are $K$ clusters. In contrast, multi-level clustering forms clusters progressively from a coarse set $Cs = \{k_1, k_2, k_3\}$. For the cluster $k_3$, high-confidence sample pairs are mined across different levels with the help of $k_1$ and $k_2$, effectively reducing the number of boundary samples and resulting in a more confident cluster structure.} 
\label{multi-resolution}
\vspace{-0.3cm}
\end{figure}

To address this issue, we propose \emph{multi-level clustering to progressively enhance the quality of cluster structures at each level}, thereby achieving more confident cluster structures. 
{Intuitively, coarse-grained classification is easier than fine-grained classification \cite{sohoni2020no}. For example, distinguishing between two broad categories (\eg animals vs. plants) is much simpler than identifying finer categories (\eg flowers, trees, chickens, dogs, and horses). Similarly, coarse-grained clustering is less prone to errors than fine-grained clustering, and thus can serve as a helpful guide for the latter.}
{As illustrated in Fig. \ref{multi-resolution} (a), clustering at a single-level may propagate boundary errors throughout the entire category, negatively impacting the final results.
In contrast, Fig. \ref{multi-resolution} (b) demonstrates that multi-level clustering gradually organizes samples into coarse-to-fine groups, selectively leveraging confident sample pairs across different levels to reduce errors. This strategy ultimately yields a more stable and accurate cluster structure.}
Besides, it is worth noting that multi-level clustering in this thesis differs from hierarchical clustering. Hierarchical clustering follows a nested structure, where the clustering assignments of the next level are based on the previous ones. In contrast, multi-level clustering conducts clustering at each level separately and then selects high-confidence sample pairs across the different levels.

{To learn a consistent and confidence cluster structure, we propose a novel method, MRG-UMC, consisting of three modules:
i) Inner-view multi-level clustering: To address instability in cluster structures caused by boundary samples and ambiguous initial clustering, we use high-confidence sample pairs across different levels for each view to establish more confidence cluster structures.
ii) Synthesized-view alignment: Direct alignment across views is challenging due to distribution discrepancies. At each level, a synthesized-view is introduced to assist in aligning other views, thereby enhancing consistency in cluster structures.
iii) Cross-view guidance: Given the different clustering performances across views, the view with the best clustering result is selected as the reliable view to guide others at each level, resulting in more confident cluster structures.}
The main contributions are as follows:
\begin{itemize}
\item {For the issue of learning consistent cluster structures with {low confidence}, we propose multi-level clustering and reliable view guidance to enhance clustering confidence.}

\item {MRG-UMC is proposed to establish consistent and confident cluster structures across views, which comprises three modules: an inner-view multi-level clustering module, a synthesized-view alignment module, and a cross-view guidance module.} 

\item {Theoretical analyses validate the multi-level clustering and reliable view guidance in boosting the cluster structure confidence. Meanwhile, extensive results on five multi-view benchmark datasets demonstrate its superior performance.}

\end{itemize}

\section{Related Work}\label{relatedwork}

\subsection{Incomplete multi-view clustering}

{Recent advancements in clustering have focused on enhancing performance and extending applications across various domains. For instance, Bacciu \etal \cite{bacciu2019bayesian} proposed a mixture model for tree-structured data clustering. Kilic \etal \cite{kilic2023binary} introduced a human-inspired method for feature selection. Additionally, clustering techniques have been successfully applied to biomonitoring \cite{duka2023approach}.}

However, in real-world scenarios, samples may be missing in one or more views due to various internal or external factors, resulting in incomplete multi-view data. To address this challenge in clustering tasks, researchers have extensively explored methods such as spectral approaches, subspace learning, and graph learning \cite{10149819, scl-UMC}. In spectral approaches, Wen \etal \cite{wen2021unified} introduced a tensor spectral clustering method to recover missing views. Besides, they utilize latent information from absent views and capture intra-view data relationships often overlooked in other methods.
Zhang \etal \cite{Unified2024Zhang} restored latent connections between observed and unobserved samples, enhancing the robustness of subspace clustering through high-level correlations. It is the first attempt to formulate incomplete multi-view kernel subspace clustering from unified and tensorized perspectives.
Wang \etal \cite{wang2023multiple} used multiple kernel-based anchor graphs to tackle IMC. which adeptly captures both intra-view and inter-view nonlinear relationships by fusing multiple complete anchor graphs across views.
To reduce the time required for recovering the original data, Zhang \etal \cite{zhang2023robust} devised an approach that integrates spectral embedding completion and discrete cluster indicator learning into a unified step, providing a solution to this issue.

For graph learning, Li \etal \cite{li2021incomplete} proposed the Joint Partition and Graph learning method, comprising two key components: unified partition space learning for noise robustness enhancement and consensus graph learning to uncover data structures.
To explore the complex relationship between samples and latent representations, Zhu \etal \cite{zhu2022latent} proposed a neighborhood constraint and a view-existence constraint for the first time.
Moreover, in cases of arbitrary missing views, He \etal \cite{he2023structured} proposed Structured Anchor-inferred Graph Learning. They replaced a fixed distance-based weighting matrix with a structural anchor-based similarity learning model, creating a trainable asymmetric intra-view similarity matrix.
Additionally, to learn the overall optimal clustering result, Xia \etal \cite{xia2023incomplete} proposed a Kernelized Graph-based IMC algorithm, which optimizes downstream sub-tasks in a mutually reinforcing manner.

For subspace learning approaches, Yin \etal \cite{2017Unified} devised a method for incomplete multi-view subspace learning, which enhances performance by simultaneously addressing feature selection and similarity preservation within and across views.
To fully leverage observed samples, Yang \etal \cite{2018Incomplete} employed a sparse and low-rank matrix to model correlations among samples within each view for imputing missing samples. They also enforce similar subspace representations to explore relationships between samples across different views, aiding clustering tasks.
Due to the coefficient matrix, the self-representation method accurately captures sample relationships. Liu \etal \cite{liu2021self} introduced a tailored self-representation subspace clustering algorithm for IMC, which integrates missing sample completion and self-representation learning into a cyclical process. Yang \etal \cite{yang2019semi} integrated the intrinsic consistency and extrinsic complementary information for prediction and cluster simultaneously.
Li \etal \cite{li2023anchor} proposed a unified sparse subspace learning framework that integrates inter-view and intra-view affinities to generate a unified clustering assignment using a sparse anchor graph. However, obtaining paired samples across all views can be challenging in real-world scenarios.

\subsection{Unpaired multi-view clustering}
UMC is an extreme scenario within IMC, where samples are solely observed in one view. This renders UMC more challenging compared to IMC \cite{10149819, scl-UMC}. 
Numerous researchers have leveraged weak supervision information to establish correlations between views. For example, Qian \etal \cite{2013Multi} and Houthuys \etal \cite{2017unpaired} integrated must-link and cannot-link constraints to build correlations across views in their studies. However, these approaches may prove ineffective without paired samples or supervised information.

For methods dealing with partially paired samples, Huang \etal \cite{huang2020partially} proposed Partially View-Aligned Clustering, which utilizes the `aligned' data to learn a common space across different views and employs a neural network to establish category-level correspondence for unaligned data. Yang \etal \cite{yang2021MvCLN} proposed Multiview Contrastive Learning with Noise-Robust Loss, which simultaneously learns representation and aligns data using a noise-robust contrastive loss. Furthermore, addressing both the partially view-unaligned and sample-missing problem, Yang \etal \cite{yang2022robust} introduced a more robust Multi-View Clustering method to directly re-align and recover samples from other views. Besides, Wang \etal \cite{wang2024partially} introduced the Cross-View Graph Contrastive Learning Network, which integrates multi-view information to align data and learn latent representations. Jin \etal \cite{jin2023deep} proposed a novel approach departing from traditional contrastive-based methods. They use pair-observed data alignment as `proxy supervised signals' to guide instance-to-instance correspondences across views. Additionally, they introduce a prototype alignment module to rectify shifted prototypes in IMVC. Zhao \etal \cite{zhao2025incomplete} learn soft clustering labels from complete data to capture category-level cross-view correspondences. However, these methods require some paired samples to establish relationships between views.

For methods dealing with unpaired data, Yang \etal \cite{zeng2023semantic} proposed the Noise-aware Image Captioning (NIC) method to make the most of trustworthy information in text with mismatched words. Specifically, NIC identified and explored mismatched words to reduce errors by assessing the reliability of word-label relationships based on inter-modal representativeness and intra-modal informativeness.
Additionally, Yang \etal \cite{10149819} introduced Iterative Multiview Subspace Learning for UMC to learn consistent subspace representations. Furthermore, they introduced another two methods to solve multi-view alignment and achieve end-to-end results, respectively. Besides, Xin \etal \cite{scl-UMC} proposed selective contrastive learning for UMC, leveraging inner-view and inter-view selective contrastive learning modules to improve cluster structure consistency. 
Furthermore, considering the differences in clustering performance across views, Xin \etal \cite{RGUMCxlk} leveraged the view with the best cluster performance to align the other views, thereby improving overall clustering performance. Although these methods explore consistent cluster structures across views, they cannot effectively mine consistency when the cluster structures are uncertain.
{Li \etal \cite{li2025scalable} address anchor and edge misalignments by proposing a Scalable Unpaired Multi-view Clustering with Bipartite Graph Matching. Notably, their definition of `unpaired’ refers to incorrect pairing relationships, whereas our work focuses on fully unpaired scenarios where no complete samples exist across any two views. That is the key distinction between them.
}

{We compared our method with prior works in four aspects:
i) Prior works does not consider unstable cluster structures, whereas MRG-UMC employs multi-level clustering to address this issue.
ii) Regarding cross-view alignment, prior works aligns only views, while MRG-UMC also aligns clusters and samples through a synthesized-view module.
iii) MRG-UMC provides theoretical support for multi-level clustering and reliable view guidance, whereas prior works lack such analysis.
iv) Extensive experiments on multiple multi-view datasets demonstrate the superior performance of our approach compared to these methods.}

{\subsection{{Dempster-Shafer evidence theory in multi-view learning}}}
There have been several studies exploring confidence with Dempster-Shafer evidence theory. The theory was first proposed by Dempster and Shafer to deal with uncertainty based on belief functions \cite{dempster2008upper, shafer1976mathematical, dempster1968generalization}. Now, it has been developed a general framework. 
Recently, Han \etal \cite{han2022multimodal} introduced multi-modal dynamics to enhance the reliability of fusion methods using evidence theory, dynamically evaluating feature-level and modality-level informativeness for dependable integration across diverse samples. 
Geng \etal \cite{geng2021uncertainty} utilize uncertainty to weigh each view of individual samples based on data quality, maximizing the use of high-quality samples while minimizing the influence of noisy ones.
Additionally, Han \etal \cite{han2022trusted} introduced a trusted multi-view classification method, dynamically integrating diverse views using the Dirichlet distribution for class probabilities and Dempster-Shafer theory for integration. Meanwhile, they also provide additional theoretical analysis to validate its effectiveness.
For unreliable predictions in long-tailed classification, Li \etal \cite{li2022trustworthy} introduced the trustworthy long-tailed classification method, which integrates classification and uncertainty estimation within a multi-expert framework, calculating evidence-based uncertainty and evidence for each expert, and merging these metrics using the Dempster-Shafer Evidence Theory.

\subsection{Multi-level feature learning for multi-view clustering}
For the method with multi-level feature learning, Xu \etal \cite{xu2022multi} proposed a method of multi-level feature learning with low-level features, high-level features, and semantic features to weaken the conflict between learning consistent semantics and inconsistent view-private information. 
To develop a robust network, Yang \etal \cite{yang2018deep} introduced the Cascade Deep Multi-Modal network structure with a cascade architecture, which maximizes correlations between layers sharing similar characteristics. Furthermore, Yang  \etal \cite{zhu2020sequential} considered both user intentions and preferences, capturing users' dynamic information in hierarchical representations to solve the next-item recommendation.
For the neglect of redundant information between views, Cui \etal \cite{cui2024novel} proposed a novel multi-view clustering framework based on information theory, utilizing the variational analysis to generate consistent information and enhance consistency through a sufficient representation of the lower bound.
Besides, to tackle the issue of semantic consistency oversight in most methods, Zhou \etal \cite{zhou2024mcoco} introduced a multi-level consistency collaborative learning framework, which learns cluster assignments of multiple views in the feature space and aligns semantic labels of different views through contrastive learning. Although these methods utilize multi-level representations or features to learn consistent information, they fail to account for the instability across different levels, which is a key distinction of our method.

\section{Methodology}\label{proposedmethod}

\subsection{Notations and Architecture}

\textbf{Notations.}
For the unpaired multi-view dataset $\{\boldsymbol{X}^v\}_{v=1}^V$, there are $V$ views. The feature dimension of each view is denoted as $d^v$. For the $v$-th view, the number of samples is $n^v$. Therefore, the total number of samples in the unpaired multi-view dataset is $N$, where $N=\sum_{v=1}^V n^v$, since no paired samples exist in UMC. Specifically, in the $v$-th view, an autoencoder consists of an encoder $F^v$ and a decoder $G^v$. We feed the $v$-th view data $\boldsymbol{X}^v$ into the corresponding encoder $F^v$ to learn the latent representation $\boldsymbol{Z}^v$, \emph{i.e.}, $\boldsymbol{Z}^v = F^v(\boldsymbol{X}^v)$. 
In the inner-view multi-level clustering module, for sample $\boldsymbol{z}_i^v$, similarity sets for true positive pairs $tp_{in}(\boldsymbol{z}_i^v)$ and the true negative pairs $tn_{in}(\boldsymbol{z}_i^v)$ are constructed in the contrastive learning. 
In the synthesized-view alignment module, a matrix of cluster pairing relationships $\boldsymbol{A}$ between views is established by the Hungarian Algorithm. 
Then, the similarity sets $pos_{cro}(\boldsymbol{z}_i^v)$ and $neg_{cro}(\boldsymbol{z}_i^v)$ for the positive and negative pairs of $\boldsymbol{z}_i^v$ are constructed. 
In the cross-view guidance module, the silhouette coefficient $sils^v$ is used to select reliable views. The set of reliable views $\mathcal{R}$ includes those with a higher silhouette coefficient than the current view. Then we employ the KL divergence (\emph{e.g.}, $D(\boldsymbol{P}^r ||\boldsymbol{Q}^v)$) to align the distribution $\boldsymbol{Q}^v$ with $\boldsymbol{P}^r$, where $\boldsymbol{P}^r$ and $\boldsymbol{Q}^v$ denote approximate probability distributions of the $r$-th and $v$-th views.
In the theoretical analysis, as suggested by Dempster-Shafer Evidence Theory \cite{han2022trusted}, $\{b_k^r\}_{k=1}^K$ and $\{b_k^o\}_{k=1}^K$ denote the belief masses from the reliable view and the original view, respectively. $l$ represents the index of ground truth, while $m$ denotes the index of the largest original belief mass in $\{b_k^o\}_{k=1}^K$. $u^r$ and $u^o$ represent the uncertainty masses for the reliable view and {the original view, }respectively, satisfying $\sum_{k=1}^K b_k^r + u^r = 1$ and $\sum_{k=1}^K b_k^o + u^o = 1$. For clarification, the main symbols and their definitions are listed in the {supplementary}. 

\begin{figure*}[!t]
\centering
\includegraphics[width=1.95\columnwidth]{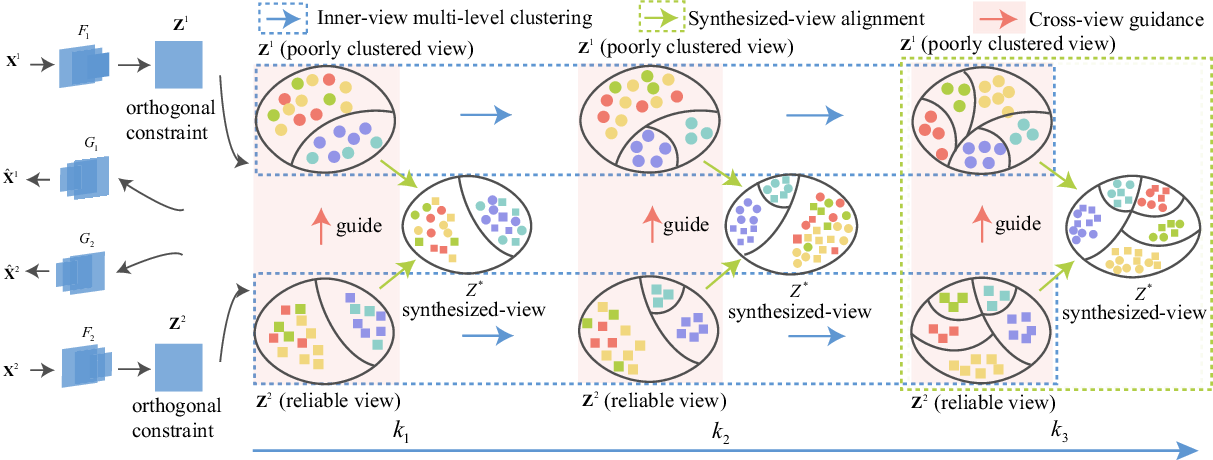}
\vspace{-0.2cm}
\caption{The framework of Multi-level Reliable Guidance for UMC is built upon autoencoders with regularization. In latent spaces, training occurs in three stages with different numbers of clusters ($k_1$, $k_2$, and $k_3$). At each stage, three modules are employed: the inner-view multi-level clustering module, the {synthesized-view alignment module}, and the cross-view guidance module.} 
\label{super-class-framework}
\vspace{-0.5cm}
\end{figure*}

{\textbf{The architecture of MRG-UMC.}}
We illustrate the training process with two views and five clusters. The cluster set $Cs = \{k_1, k_2, k_3\}$ is defined as $\{2, \lceil K/2 \rceil, K\}$, where $K$ is the true number of clusters, similar to \cite{gui2022improving}.
As depicted in Fig. \ref{super-class-framework}, the two-view samples ($\boldsymbol{X}^1, \boldsymbol{X}^2$) are fed into two autoencoders with regularization to learn view-specific subspace representations ($\boldsymbol{Z}^1, \boldsymbol{Z}^2$). 
Subsequently, in the subspace, three modules are used to learn consistent and confident cluster structures. Initially, the clustering begins with $k_1$ clusters, creating coarse-grained groupings. As training goes, the number of clusters increases to $k_2$, leading to finer-grained clusters. In the last stage, the cluster count is adjusted to $k_3$, resulting in a more refined cluster structure with the true number of clusters. 
For example, when the cluster number is $k_3$, in the inner-view multi-view clustering (indicated by the blue dashed box), a more confident cluster structure is established by leveraging confident sample pairs across three levels. 
In the synthesized-view alignment module (depicted in the green dashed box), each view aligns with the synthesized-view to reduce distribution discrepancies, thereby helping to learn consistent cluster structures.
Furthermore, in the cross-view guidance module (shown in the red rectangular box), the reliable view guides the poorly clustered view to form a more confident cluster structure.
After convergence, we perform $K$-means clustering on the concatenated matrix $\boldsymbol{Z}^* = [\boldsymbol{Z}^1; \boldsymbol{Z}^2]$ to obtain the final clustering assignments.

\textbf{Multi-view autoencoders with regularizer.}
Autoencoder \cite{xu2022multi} serves as a prevalent unsupervised model for projecting raw features into a latent feature space. Typically, an autoencoder includes an encoder and a decoder. 
For a given feature representation $\boldsymbol{X}^v$ in the \emph{v}-th view, we use an autoencoder to derive the latent representation $\boldsymbol{Z}^v$ by minimizing the reconstruction loss. Additionally, to enhance the discriminative of $\boldsymbol{Z}^v$, we apply an orthogonal constraint to prevent it from expanding arbitrarily within the feature space \cite{chen2022adaptively} \cite{chen2022efficient}. Therefore, for multi-view data, we obtain latent representations using autoencoders with a regularizer \cite{10149819,scl-UMC} by:
\begin{equation}\label{eq1}
\small
l_{AE} = \sum_{v=1}^V ||\boldsymbol{X}^v - G^v\big(F^v(\boldsymbol{X}^v)\big)||_F^2 + \lambda_1 ||{\boldsymbol{Z}^v}^{T}\boldsymbol{Z}^v - \boldsymbol{I}_{d}||_F^2,
\\
\end{equation}
where $F^v$ and $G^v$ represent the encoder and decoder of the \emph{v}-th autoencoder, respectively. $\boldsymbol{Z}^v$ denotes the latent representation of $\boldsymbol{X}^v$. $\lambda_1$ is a hyperparameter that adjusts the balance between the reconstruction term and the orthogonal constraint.

\subsection{Methodology}
\textbf{Multi-level clustering.}
Due to the lack of supervision in clustering, cluster assignments can be uncertain, especially for boundary samples. To mitigate the influence of such samples, various strategies have been proposed, including progressive sampling \cite{yang2024not, Yang2024Robust} and adaptive weighting \cite{shen2023depression, liu2023adaptive, he2022not}.
Typically, boundary samples are assigned small weights to reduce their influence. However, identifying boundary samples using a single-level approach may be unreliable.
In addition, initial cluster assignments may be unstable due to limited training. To address these issues, we adopt a multi-level clustering strategy that categorizes samples progressively, thereby alleviating the negative effects of boundary samples and unstable initial assignments.
Specifically, we draw on the concept of super-class \cite{gui2022improving, zhang2021flexmatch, sohn2020fixmatch} to construct the multi-level clustering.
At each level, the method encourages samples to stay closer to the distribution of their corresponding super-class rather than those of others.
Compared to direct pseudo-label assignment through single-level clustering, multi-level clustering offers softer results to `roughly categorize' samples and reduces clustering errors step by step.

\textbf{Inner-view multi-level clustering.}
In each view, samples can be grouped into clusters based on their distance from various cluster centroids. However, without supervised information, clustering assignments remain inherently uncertain. 
Contrastive learning is an unsupervised representation learning method that improves cluster performance by maximizing the similarities of positive pairs and minimizing the similarities of negative pairs in the feature space \cite{xu2022multi}. 
In clustering, the more accurately the relationships between samples are captured, the more confidence the cluster structures can be characterized.
Therefore, to achieve confident cluster structures, instead of selecting high-confidence samples in contrastive learning, we leverage \textbf{high-confidence sample pairs}. 

Following scl-UMC \cite{scl-UMC}, we treat samples within the same cluster as positive pairs and those from different clusters as negative pairs. We then refine these pairings by classifying them into true and false categories based on the consistency of cluster relationships across multiple levels.
For positive pairs, those that retain their positive correlation at each level are considered true positives, while those whose relationships vary across levels are labeled as false positives.
Similarly, negative pairs that consistently show negative correlations across levels are classified as true negatives, whereas pairs with inconsistent correlations are designated as false negatives.

Therefore, to enhance cluster structure confidence, high-confidence sample pairs, consisting of true positive and true negative examples, are employed in contrastive learning. To establish the correlation between samples, we first calculate the similarity between samples $\boldsymbol{z}^v_i$ and $\boldsymbol{z}^v_j$ using cosine similarity:
\begin{equation}\label{cosine}
s^{v}_{ij}= \frac{\boldsymbol{z}^{v}_i \boldsymbol{z}^{{v}^{\top}}_j} {||\boldsymbol{z}^{v}_i||\cdot||\boldsymbol{z}^{v}_j||}. 
\end{equation}
Similarly, we define the super-class set as \cite{gui2022improving}, \eg {$Cs = \{k_1, k_2, k_3\}$}.
Then, the similarities of true positive pairs $tp_{in}(\boldsymbol{z}_i^{v})$ and true negative pairs $tn_{in}(\boldsymbol{z}_i^{v})$ for $\boldsymbol{z}_i^{v}$ in the inner-view across three levels are designed as follows: 
\begin{equation}\label{eq11}
\begin{split}
     tp_{in}(\boldsymbol{z}_i^{v}) =&\{s_{ij}^{v}: j=1,2,\ldots, m_{i}^{v}, j \neq i, \\
    & ks({\boldsymbol{z}^{v,k_1}_i}) \cap 
    ks({\boldsymbol{z}^{v,k_2}_i})\cap ks({\boldsymbol{z}^{v,k_3}_i}) \},\\        
     tn_{in}(\boldsymbol{z}_i^{v}) =&\{\widetilde{s_{ij}^{v}}: j=1,2,\ldots, n_{i}^{v}, j \neq i,\\
    & kd({\boldsymbol{z}^{v,k_1}_i})  \cap  kd({\boldsymbol{z}^{v,k_2}_i})  \cap 
    kd({\boldsymbol{z}^{v,k_3}_i})\},\\ 
\end{split}
\end{equation}
where $m_{i}^{v}$ and $n_{i}^{v}$ denote the number of true positive pairs and true negative pairs of $\boldsymbol{z}_i^{v}$, respectively, intersecting across three different cluster levels. 
$ks(\boldsymbol{z}_i^{v,k_1})$ and $kd(\boldsymbol{z}_i^{v,k_1})$ denote the sets of samples with the same and different cluster assignments as $\boldsymbol{z}_i^{v}$ for $k_1$ clusters, respectively. Similarly, $ks(\boldsymbol{z}_i^{v,k_2})$ and $kd(\boldsymbol{z}_i^{v,k_2})$ indicate the sets for $k_2$ clusters, while $ks(\boldsymbol{z}_i^{v,k_3})$ and $kd(\boldsymbol{z}_i^{v,k_3})$ refer to the sets for $k_3$ clusters.

With the assistance of high-confidence sample pairs, we then use the NT-Xent contrastive loss \cite{chen2020simple} across all views to learn the confident cluster structures across views as follows:
\begin{equation}\label{eq9}
l_{in} = \frac{1}{V}\sum_{v=1}^V \sum_{i=1}^{n^v} \sum_{j=1}^{m_i^v} \frac{1}{n^v m_i^v} {l}_{ij}^{v},
\end{equation}
where
\begin{equation}\label{eq10}
l_{ij}^{v}  = -\log \frac{\exp{(s_{ij}^{v}/\tau)}}{\sum_{\widetilde{s_{ij}^{v}}\in tn_{in}(\boldsymbol{z}_i^{v})} \exp({\widetilde{s_{ij}^{v}}/\tau})}.
\end{equation}
$n^v$ is the number of samples in the $v$-th view, and $m^v_i$ represents the number of true positive pairs in the similarity set of $\boldsymbol{z}^{v}_i$, \emph{i.e.}, $m^v_i = |tp_{in}(\boldsymbol{z}_i^{v})|$. The set $tn_{in}(\boldsymbol{z}_i^{v})$ contains the true negative pairs of $\boldsymbol{z}_i^{v}$. Besides, the temperature parameter $\tau$ is set to 0.1.

\textbf{Synthesized-view alignment.}
Each view has its specific feature representation, making it challenging to directly align. To address this, we introduce a synthesized-view as an auxiliary view to reduce the discrepancy cross views.
Specifically, the synthesized-view is constructed by concatenating multiple individual views, \emph{i.e.}, $\boldsymbol{Z}^* = [\boldsymbol{Z}^1; \ldots; \boldsymbol{Z}^V]$. 
Furthermore, to learn consistent cluster structures between views, we leverage contrastive learning to align each single view $\{\boldsymbol{Z}^v\}_{v=1}^{V}$ with the synthesized-view $\boldsymbol{Z}^*$.
However, without paired samples, it is challenging to explore the relationships between views. Therefore, we turn to establishing cluster pairing relationships and then aligning these clusters across views.

Firstly, we aim to construct the cluster pairing relationships between views. Specifically, we obtain the cluster centroids $\boldsymbol{C}^*$ and $\boldsymbol{C}^v$ from the synthesized-view and the $v$-th view using $K$-means clustering. Then, we calculate the weight matrix $M^{v}$ using $\frac{\boldsymbol{C}^*(\boldsymbol{C}^v)^{\top}}{||\boldsymbol{C}^*|| \cdot ||(\boldsymbol{C}^v)^{\top}||}$.
{To acquire the pairing relationship, we transform the task of finding pairs of cluster centroids from two views into maximum-weight matching in a bipartite graph \cite{grinman2015hungarian, bruff2005assignment}.}
{The objective is to choose K entries from $\boldsymbol{M}^{v}$, where entries from each row and column maximize the overall weight matrix. Thus, learning the centroids pairing relationship becomes a maximum-weight matching issue, which can be solved by the Hungarian algorithm \cite{wright1990speeding}.}
Naturally, the pairing relationship of cluster centroids $\boldsymbol{A}^{v}$ between the synthesized-view $\boldsymbol{Z}^*$ and the $v$-th view $\boldsymbol{Z}^v$ is solved by the Hungarian Algorithm \cite{scl-UMC} as follows:
\begin{equation}\label{eq12}
\begin{split}
& {\max_{\boldsymbol{A}^{v}} \sum_{i=1}^K \sum_{j=1}^K {m}^{v}_{ij} {a}^{v}_{ij},}\\
s.t. \boldsymbol{A}^{v}& {\boldsymbol{A}^{v}}^{\top} = \mathbf{I}_{K}, \boldsymbol{A}^{v} \in \{0,1\}^{K \times K}\\
\end{split}
\end{equation}
where ${m}_{ij}^{v} \in \boldsymbol{M}^{v}$, ${a}^{v}_{ij} \in \boldsymbol{A}^{v}$. Then, the pairing correlation of clusters across views is obtained.

Secondly, we use contrastive learning to align pairing clusters. Specifically, in contrastive learning, positive sample pairs are samples with pairing clusters between a single view and the synthesized-view, while negative sample pairs are those with non-pairing clusters \cite{scl-UMC}. 
Similarly to Eq.(\ref{cosine}), we calculate the similarity between $\boldsymbol{z}_i^*$ and $\boldsymbol{z}_j^v$ with cosine similarity ${s}^{*v}_{i,j}= {\boldsymbol{z}_{i}^{*}\boldsymbol{z}_j^{v^{\top}}}/({||\boldsymbol{z}_{i}^{*}||\cdot||\boldsymbol{z}_j^v||}).$
Then the positive similarity set $pos_{cro}(\boldsymbol{z}_i^*)$ and negative similarity set $neg_{cro}(\boldsymbol{z}_i^{*})$ of $\boldsymbol{z}_i^*$ are constructed as follows: 
\begin{equation}\label{eq13}
\fontsize{9}{10}
\begin{split}
    & pos_{cro}(\boldsymbol{z}_i^*) = \{s_{ij}^{*v}: v=1,2,\ldots, V, j=1,2,\ldots, n^v,  a^{v}_{ij}=1\},\\
    & neg_{cro}(\boldsymbol{z}_i^{*}) = \{\widetilde{s_{ij}^{*v}}: v=1,2,\ldots, V, j=1,2,\ldots, n^v, a^{v}_{ij}=0\},
\end{split}
\end{equation}
where $s_{ij}^{*v}$ and $\widetilde{s_{ij}^{*v}}$ are the similarities of $\boldsymbol{z}_i^*$ and $\boldsymbol{z}_j^v$ in positive pairs and negative pairs, respectively. $a^{v}_{i,j}=1$ denotes that the cluster of $\boldsymbol{z}_i^*$ in the synthesized-view matches the cluster of $\boldsymbol{z}_j^v$ in the $v$-th view, and vice versa for $a^{v}_{i,j}=0$. Therefore, the synthesized-view contrastive learning is formed as follows: 
\begin{equation}\label{eq14}
l_{sy} =  \frac{1}{N V }\sum_{i=1}^{N} \sum_{v=1}^{V} \sum_{j=1}^{n^v} \frac{1}{n^v }{l}^{*v}_{i,j},
\end{equation}
where 
\begin{equation}\label{eq15}
{l}^{*v}_{i,j} = -\log \frac{ \exp(s_{i,j}^{*v}/\tau)}{\sum_{\widetilde{s_{i,j}^{*v}}\in neg_{cro}} \exp({\widetilde{s_{i,j}^{*v}}/\tau})}.
\end{equation}
$N$ and $n^v$ represent the number of samples in the synthesized-view and the $v$-th view, respectively. The temperature parameter $\tau$ is set to 0.1.

{\textbf{Cross-view guidance.}}
Given the different clustering performances across views, we choose well-clustered views as reliable views to guide those with poor clustering performance. To assess the clustering quality of each view, we use the silhouette coefficient, which evaluates how well objects are grouped by comparing the distances within their cluster to those of other clusters \cite{lin2022tensor}.
In multi-view learning, the silhouette coefficient of the $v$-th view is defined as follows:
\begin{equation}\label{eq7}
sils^v = \frac{1}{n^v} \sum_{i=1}^{n^v} sil(\boldsymbol{z}_i^v), 
\end{equation}
where $sil(\boldsymbol{z}_i^v)$ is the silhouette coefficient of sample $\boldsymbol{z}_i^v$ \cite{lin2022tensor}. 
As for guidance, we employ the KL divergence \cite{RGUMCxlk} (\emph{e.g.}, $D(\boldsymbol{P}^r ||\boldsymbol{Q}^v)$), an asymmetric metric, to guide the distribution $\boldsymbol{Q}^v$ with $\boldsymbol{P}^r$, where $\boldsymbol{P}^r$ and $\boldsymbol{Q}^v$ denote approximate probability distributions of the $r$-th (reliable view) and $v$-th view.

However, in the early stages of training, the clusters are not well-formed, leading to a low silhouette coefficient. Consequently, only the most reliable view is suitable for guiding the current views. As training goes on and the cluster structure becomes clearer, more views can be considered reliable for guiding the current view. To capture this process, we use a decreasing coefficient $\tau$ to dynamically select reliable views. 
Specifically, the initial coefficient $\tau$ is set to 1.5 and decreases by 0.99 each epoch until it reaches 1. For each view, the set of reliable views is defined as those with a silhouette coefficient greater than $\tau$ times that of the current view, denoted as $\mathcal{R}$. 
Consequently, the loss for the cross-view guidance module is formulated as follows:
\begin{equation}\label{eq10-2}
\setlength\abovedisplayskip{1.5pt}
\setlength\belowdisplayskip{1.0pt}
l_{cr} = \sum_{v=1}^V \sum_{r=1}^{|\mathcal{R}|} \frac{1}{V^2} \boldsymbol{P}^r \text{log}(\frac{\boldsymbol{P}^r}{\boldsymbol{Q}^v}),
\end{equation}
where the indices of the reliable view and the current view are denoted by $r$ and $v$, respectively. $|\mathcal{R}|$ represents the number of reliable views for the $v$-th view.

In summary, MRG-UMC is formulated to capture consistent and confident cluster structures as follows:
\begin{equation}\label{eq13-1}
L = l_{AE}+ \lambda_2 l_{in} +\lambda_3 l_{sy}+ \lambda_4 l_{cr} ,\\
\end{equation}
where $l_{AE}$, $l_{in}$, $l_{sy}$ and $l_{cr} $ are the terms of an autoencoder network with orthogonal constraint, inner-view multi-level clustering, synthesized-view alignment, and cross-view guidance, respectively. Besides, $\lambda_2$, $\lambda_3$, and $\lambda_4$ are the hyperparameters to balance these terms.

\textbf{Algorithm and convergence analysis.}
The algorithm for MRG-UMC and the training loss curves for four datasets are provided in the supplementary.
{We further analyze the convergence. Each term in Eq. (\ref{eq13-1}) is non-negative. Hence, the total objective function is bounded below by zero, \ie $L \geq 0$. In addition, $\mathcal{L}$ is smooth and differentiable almost everywhere. During training, we employ the Adam optimizer with a fixed learning rate of 0.0001, a mini-batch size of 256, and default hyperparameters. According to theoretical results in \cite{bottou2018optimization}, Adam converges to a first-order stationary point under the following mild conditions:
\begin{itemize}
    \item The objective function is smooth and bounded below;
    \item The variance of the stochastic gradients is finite;
    \item The learning rate is sufficiently small and constant.
\end{itemize}
Under these conditions, the training loss is almost surely guaranteed to converge to a local minimum \ie $\lim_{t \to \infty} \mathbb{E}\left[\|\nabla \mathcal{L}(\theta_t)\|^2\right] = 0$. More detailed theoretical discussions are provided in the supplementary material.}

{In our experiments, we empirically define convergence as the point at which the relative change in loss, $|L_t - L_{t-1}| / L_{t-1}$, remains below $1 \times 10^{-4}$ for 10 consecutive epochs, within a maximum of 200 epochs. As shown in Figure 1 (supplementary), our model consistently satisfies this criterion across all datasets. These empirical results, along with the satisfaction of the above theoretical conditions, support the convergence of our method.}

\vspace{-0.3cm}
\begin{figure}
    \centering    
    \includegraphics[width=1\columnwidth]{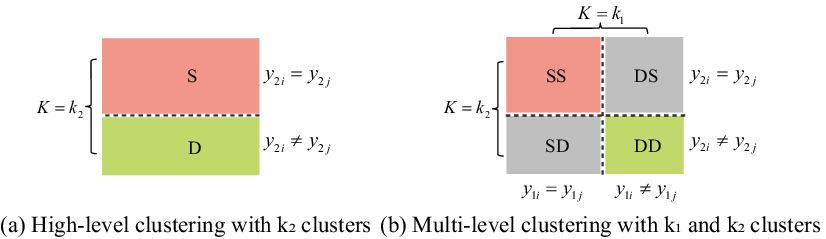}
    \vspace{-0.5cm}
    \caption{The Venn diagram illustrates the sample sets in high-level and multi-level clustering, where $k_1 < k_2$. `S' denotes the similar set, and `D' denotes the dissimilar set.}
    \label{Venn_diagram}
    \vspace{-0.5cm}
\end{figure}

\subsection{{Theory Analysis}}
To enhance the confidence of cluster structure, we employ two strategies: multi-level clustering and reliable view guidance.
Then, we theoretically validate the effectiveness of these strategies and arrive at the following conclusions: i) Multi-level clustering helps reduce inconsistencies across different levels, thereby improving the confidence of cluster structures. ii) Guidance from a reliable view can potentially enhance the belief masses of poorly clustered views, thereby increasing the confidence in these views.

\textbf{Proposition 1.} 
\emph{Given a dataset $\boldsymbol{X}$, assuming it initially yields a low-level cluster assignment $\boldsymbol{Y_1}$ with $k_1$ clusters, followed by a high-level cluster assignment $\boldsymbol{Y_2}$ with $k_2$ clusters ($k_1 < k_2$). Then four sets of samples are generated based on the cluster assignments:}
\begin{equation}\label{sample_relationship2}
    \begin{split}
        SS &= \{(x_i,x_j)|y_{1i} = y_{1j}, y_{2i} = y_{2j},i \neq j\},\\
        DD &= \{(x_i,x_j)|y_{1i} \neq y_{1j},y_{2i} \neq y_{2j},i \neq j\},\\
        SD &= \{(x_i,x_j)|y_{1i} = y_{1j},y_{2i} \neq y_{2j},i \neq j\},\\
        DS &= \{(x_i,x_j)|y_{1i} \neq y_{1j},y_{2i} = y_{2j},i \neq j\}.
    \end{split}
    \vspace{-0.4cm}
\end{equation}
{\emph{In multi-level clustering with high-level and low-level clusters, removing sample sets (SD and DS) with inconsistent relationships between the two levels reduces cluster structure inconsistency, thereby improving the confidence in the cluster structure.}}

As shown in Fig. \ref{Venn_diagram} (a), samples are directly divided into a similarity set ($S$) and a dissimilarity set ($D$) in high-level clustering with $k_2$ clusters. However, clustering errors are inevitable. In Fig. \ref{Venn_diagram} (b), in high-level clustering with cluster assignment $Y_2$, removing sample sets $SD$ with different cluster relationships in $Y_1$ would reduce cluster structure inconsistency and enhance the confidence of the high-level cluster structure. Besides, the $DS$ set contains few samples, and even if they exist, they are likely boundary samples. Removing the $DS$ set will certainly reduce clustering inconsistency. Therefore, multi-level clustering is better than single-level clustering for learning the confident cluster structure.

\textbf{Proposition 2.} 
\emph{Based on Dempster-Shafer Evidence Theory \cite{han2022trusted, wang2022clustering}, guidance from reliable views enhances the belief mass in the original view.
{That is, for the $l$-th class (the ground-truth class), under the condition that the belief mass of the reliable view ($b_l^r$) is not lower than any belief mass from the original view ($\{b^o_k\}_{k=1}^K$), the reliable view can enhance the belief mass of the original view ($b_l \geq b_l^o$). Here, $b_l^o$ and $b_l$ represent the belief masses of the original view for the ground-truth class before and after integrating the reliable view, respectively.}}

\textbf{Proof.}
\begin{equation}
\small
\begin{split}
    b_l 
     & = \frac{b^o_l b^r_l + b^o_l u^r + b^r_l u^o}{\sum _{k=1}^K b^o_k b^r_k + u^o + u^r - u^o u^r} \quad  \text{{//} \textcolor{lightgray}{Eq.(13) in \cite{han2022trusted}}}\\
     & \geq \frac{b^o_l (b^r_l+u^r+u^o)}{b_m^o(1-u^r) + u^o + u^r - u^o u^r} \quad \text{{//} \textcolor{lightgray}{$b_l^r \geq b_l^o$}}\\ 
     & \geq \frac{b^o_l (b^r_l+u^r+u^o)}{b_m^o + u^r + u^o} \quad \quad \text{{//} \textcolor{lightgray}{Removing $(-b_m^o u^r-u^o u^r)$}}\\
     & \geq b_l^o. \quad \text{{//} \textcolor{lightgray}{$b_l^r \geq b_m^o$}}
\end{split}
\end{equation} 
{where $b_m^o$ denotes the largest original belief mass in the set $\{b_k^o\}_{k=1}^K$.} 
Therefore, for the $l$-class, if the belief mass $b_l^r$ on the reliable view exceeds any value in the original set of belief masses $\{b_k^o\}_{k=1}^K$, integrating $b_l^r$ from the reliable view can potentially enhance the belief mass $b_l^o$ of the original view, thereby increasing confidence in the cluster structure. A more detailed analysis is provided in the supplementary material.

{Clustering is inherently unsupervised and sensitive to boundary samples and initialization, often resulting in cluster structures with low confidence. To address this, our theoretical analysis is designed to enhance confidence in clustering structures. Besides, these two propositions have wide applicability.
Specifically, Proposition 1 supports a more confident cluster structure by reducing the influence of boundary samples, which is crucial for improving stability in applications such as autonomous driving and medical diagnostics. Proposition 2 demonstrates that leveraging reliable views to guide others can significantly improve the belief mass of the other view, especially in real-world settings like cross-modal healthcare analysis and financial fraud detection.}

\begin{table*}
\centering
\vspace{-1mm}
\renewcommand{\arraystretch}{1.1}
\setlength{\tabcolsep}{4.5pt}
\caption{Comparison of related multi-view clustering methods on \textbf{multi-view} datasets. The symbol '-' indicates out of memory.}\label{table5_multi_view_morethantwo} 
\begin{tabular}{l|ccc|ccc|ccc|ccc|ccc}
\hline\hline
\multirow{2}{*}{\textbf{Methods}}& \multicolumn{3}{c}{\emph{Digit (6views)}} \vline &\multicolumn{3}{c}{\emph{Scene-15 (3views)}} \vline &\multicolumn{3}{c}{\emph{Caltech101-20 (6views)}} \vline
&\multicolumn{3}{c}{\emph{Flower17 (7views)}}\vline&\multicolumn{3}{c}{\emph{Reuters (5views)}}\\ \cline{2-16}
&\emph{NMI} &\emph{ACC} &\emph{F1} &\emph{NMI} &\emph{ACC} &\emph{F1} &\emph{NMI} &\emph{ACC} &\emph{F1} &\emph{NMI} &\emph{ACC} &\emph{F1}
&\emph{NMI} &\emph{ACC} &\emph{F1}\\\hline

\textbf{OPMC} {(ICCV2021)}  &12.31&18.33&17.42   &17.23&18.94&14.01  &12.42&14.30&14.69  &13.52&13.38&10.14  &11.11&19.48&20.41 \\
\textbf{OP-LFMVC} {(ICML2021)}
&22.54&11.57&19.90  &9.73&17.02&11.02   &13.38&17.34&13.56  &9.71&14.88&8.02    &-&-&-  \\ 
\textbf{DAU-Nets} {(AAAI2021)} &17.10&19.40&16.82    &22.62&21.52&16.30  &24.50&15.34&17.90  &14.84&13.81&10.77  &2.20	&25.06	&22.98 \\
\textbf{RMVC} {(AAAI2018)} &22.31	&26.25	&17.88    &28.84 &24.68 &	18.03 &21.72 &18.90 &17.34 &19.08 &17.28	&11.24 &- &- &-\\

\textbf{DSMVC} {(CVPR2022)}& 13.95	&16.50	&7.80 & 19.86	&19.38 &10.09 &19.44 &14.71 &2.60 &13.55 & 12.50	&4.57 &1.53 &19.06 &15.12\\

\textbf{AWMVC} {(AAAI2023)} &12.72 &	18.65	&14.79 &20.95	&22.08	&15.08 &22.32	&17.23	&18.10 &14.17	&13.68	&9.04 &2.57	&23.44	&24.01\\
{\textbf{UOMvSC} {(TKDE2023)}} &15.51	&18.85	&18.17 &26.59	&24.55	&16.75 &24.48	&16.64	&17.50 &14.70	&13.09	&10.99  &-&-&- \\
{\textbf{WMVEC-FP}} {(KBS2024)} &10.75	&18.25	&16.24 &7.34	&31.56	&26.23
&17.19	&19.04	&12.09 &1.92	&6.62	&10.89 &3.77	&28.37	&32.78\\
{ \textbf{CVCL}} {(ICCV2023)}   &67.35	&61.70	&55.69    &38.37	&33.33	&31.88    &48.86	&31.82	&24.35    &45.64	&44.58	&38.39 &- &- &- \\\hline

\textbf{DAIMC} {(IJCAI2018)}
&18.17&25.68&17.96  &12.49&19.46&18.68  &12.49&19.46&18.68  &8.56&12.75&9.83     &-&-&- \\
\textbf{UEAF} {(AAAI2019)}
&10.32&19.25&16.63  &11.59&15.65&12.56  &9.28&27.03&24.26   &3.91&7.54&10.74    &-&-&- \\
\textbf{IMSC-AGL} {(TCYB2020)}
&24.01&17.61&17.38  &16.45&20.88&12.99  &17.30&15.81&12.97  &12.66&14.93&8.69    &-&-&- \\
\textbf{OMVC} {(Bigdata2017)}
&12.02&20.03&16.70  &11.53&16.09&11.74  &18.29&19.26&17.31  &12.68&14.68&9.13    &-&-&- \\
\textbf{OPIMC} {(AAAI2019)}  &20.13&27.74&19.66  &15.48&20.33&13.35  &17.10&20.18&16.71  &11.76&16.39&8.87   &10.92&27.46&26.13 \\
\textbf{FCMVC} {(TIP2024)} &12.39	&22.55	&14.91 &9.32	&17.15	&10.17 &9.73	&13.29	&10.28
&8.05	&14.59	&7.46 &5.46	&24.93	&21.74\\
\textbf{T-UMC} {(TCYB2022)} &9.09	&17.70	&15.06
&26.51	&27.65	&22.91 &11.95	&14.08	&9.01 &11.66	&15.74 &13.57 &-&-&-\\
{\textbf{GIGA}} {(PR2024)} &8.84	&17.50	&18.95 &15.45	&16.48	&15.67 &21.16	&16.26	&13.03 &15.73	&15.00	&15.40 &-&-&-\\\hline
{\textbf{UPMGC-SM}} {(AAAI2024)} &31.43&35.36&28.47 &21.35&26.04&16.50 &16.84&16.49&10.58 &26.20&24.65&16.91  &-&-&- \\

\textbf{IUMC-CA} {(TNNLS2023)}
&75.16&54.53&60.82    &\underline{50.25}&{34.98}&34.47 &58.31&33.42&33.24  &\textbf{66.13}&{48.52}&{44.19}   &-&-&- \\ 
\textbf{IUMC-CY} {(TNNLS2023)}
&{74.64}&{60.72}&{61.17}   &\textbf{56.75}&30.58&{35.34} &{61.48}&{46.63}&\bf{45.06}   &{63.23}&{49.21}&40.19  &10.07&22.99&21.48 \\ 
\textbf{scl-UMC} {(TNNLS2023)} &{72.75}	&{60.00}	&{50.97} &49.62	&39.15	&33.51    &59.63	&57.94	&{60.93}    &\underline{65.65}	&49.93	&47.80  &30.73&52.83&31.32\\
\textbf{RG-UMC} {(TNNLS2024)} &{77.37} &{84.85}&{84.65}  &{49.23}	&\underline{53.82}	&\underline{49.82}  &{65.81}&{67.05}&{38.92}      &58.57&{50.04}&{47.81}	&{30.93}&\textbf{59.78}&\underline{34.54} \\
\textbf{RGs-UMC} {(TNNLS2024)} &\underline{86.19}&\underline{92.83}&\underline{92.77}       &47.40&{53.50}&{48.11}    &\underline{70.88}&\underline{75.28}&{41.50}     &{60.64}&\underline{52.02}&\underline{51.68} &\underline{32.26}&{57.24}&{33.35}\\
\textbf{MRG-UMC} &\textbf{88.36}	&\textbf{93.99}	&\textbf{93.92}	&49.42&\bf{53.88}&\bf{50.47} &\textbf{72.82}	&\textbf{76.48}	&\underline{42.01} &{61.76}	&\textbf{54.85}	&\textbf{53.33}	&\textbf{40.71}	&\underline{58.29}	&\textbf{37.47}\\		
\hline\hline
\end{tabular}
\vspace{-0.5cm}
\end{table*}

{\section{TESTS}}\label{Tests}
{In this section, we conduct a series of tests to evaluate the effectiveness of MRG-UMC, including comparative experiments, verification experiments for two key propositions, ablation studies, visualization and so on.} {Besides, the hyperparameter analysis is provided in the supplementary material.}

\footnotetext[1]{http://archive.ics.uci.edu/dataset/72/multiple+features}
\footnotetext[2]{https://github.com/XLearning-SCU/2021-CVPR-Completer/tree/main/data}
\footnotetext[3]{http://www.robots.ox.ac.uk/vgg/data/flowers/17/index.html}
\footnotetext[4]{http://archive.ics.uci.edu/ml/datasets/Reuters+RCV1+RCV2+Multilingual\\\%2C+Multiview+Text+Categorization+Test+collection}

\textbf{Datasets.} The five benchmark multi-view datasets (\emph{Digit}\footnotemark[1], \emph{Scene-15}\footnotemark[2], \emph{Caltech101-20}\footnotemark[2], \emph{Flower17}\footnotemark[3], and \emph{Reuters}\footnotemark[4]) used in this paper are the same as those in \cite{scl-UMC}. To simulate the unpaired multi-view scenario, samples from each view of these multi-view datasets are randomly selected at a ratio of $1/V$ ($V$ is the number of views), ensuring no paired samples between views. Specifically, to maintain class balance, $1/V$ of the samples from each class are selected in each view, forming the unpaired multi-view data.

\textbf{Comparison methods.} Among the 28 state-of-the-art comparison methods, we include 9 complete MC methods (OPMC \cite{liu2021onelarge}, OP-LFMVC \cite{liu2021one}, DUA-Nets \cite{geng2021uncertainty}, {RMVC}\cite{tao2018reliable}, {DSMVC} \cite{tang2022deep}, {AWMVC} \cite{wan2023auto}, {UOMvSC} \cite{Tang2023Unified}, {WMVEC-FP} \cite{liu2024adaptive} and {CVCL} \cite{chen2023deep}), 11 IMC methods (DAIMC \cite{2018Doubly}, UEAF \cite{Jie2019Unified}, IMSC-AGL \cite{2020Jie}, OMVC \cite{Shao2017Online}, OPIMC \cite{2019One}, MvCLN \cite{yang2021MvCLN}, Completer \cite{lin2021completer}, {FCMVC \cite{Wan2024Fast}}, T-UMC \cite{lin2022tensor}, GIGA \cite{yang2024geometric}, and {ICMVC} \cite{chao2024incomplete}), and 8 UMC methods ({UPMGC-SM} \cite{wen2023unpaired}, IUMC-CA \cite{10149819}, IUMC-CY \cite{10149819}, MGCCFF \cite{zhao2025incomplete}, {SURE} \cite{yang2022robust}, scl-UMC \cite{scl-UMC}, RG-UMC \cite{RGUMCxlk}, and RGs-UMC \cite{RGUMCxlk}). 
All complete and incomplete multi-view learning methods adopt mean imputation to handle missing data, consistent with their original experimental settings. Moreover, all comparative methods are implemented using open-source code with the optimal parameters recommended in their respective papers.

\textbf{Implementation details.} {MRG-MUC is implemented in PyTorch 1.13, and all evaluations run on a standard Ubuntu 20.04 OS with NVIDIA 2080Ti GPU.} The network architecture used in our work follows that of \cite{lin2021completer,scl-UMC}, with each layer including a batch normalization layer and a ReLU layer. {For training, we set the maximum number of epochs to 200 and employed mini-batch gradient descent with a batch size of 256. We used the Adam optimizer with default parameters to train our model, and the learning rate is set to 0.0001.}
Specifically, the cluster set $Cs$ is $\{2, \lceil K/2 \rceil, K\}$ for all datasets, as in \cite{gui2022improving}, where $K$ denotes the ture number of clusters.
{Once the training process converges, all unpaired samples are passed through the autoencoder network with the saved parameters to obtain the latent feature representations $\{\boldsymbol{Z}^v\}_{v=1}^V$. We then obtain the final clustering result by applying $K$-means clustering to the concatenated representation $\boldsymbol{Z}=[\boldsymbol{Z}^1; \boldsymbol{Z}^2; \ldots; \boldsymbol{Z}^V]$.} We use accuracy (ACC), Normalized Mutual Information (NMI), and F1-score (F1) as evaluation metrics and report the average results from 10 runs.

\subsection{Performance Comparison and Analysis}

\textbf{Clustering performance comparison of methods on multi-view datasets.}
{We compared our model with 23 methods across five multi-view datasets, excluding MvCLN \cite{yang2021MvCLN}, Completer \cite{lin2021completer}, ICMVC \cite{chao2024incomplete}, {MGCCFF \cite{zhao2025incomplete}}, and SURE \cite{yang2022robust}, which are specifically designed for two-view settings.} From TABLE \ref{table5_multi_view_morethantwo}, we observe the following:

i) Our method surpasses all MC and IMC methods, demonstrating its superior effectiveness in UMC over the MC and IMC approaches. Typically, MC and IMC methods rely on paired samples between views for clustering. However, when handling unpaired multi-view data, missing information must be recovered, as noted in \cite{Wan2024Fast}, which is inherently prone to errors.
Fortunately, our method constructs relationships between views through pairing clusters, avoiding recovery errors. 

ii) Compared with six state-of-the-art UMC methods, our model achieves average improvements of {12.95\%, 14.99\%, and 11.62\%} in NMI, ACC, and F-score, respectively, indicating a positive effect on clustering performance. Specifically, MRG-UMC achieves an average improvement of 3.14\% in NMI, 1.32\% in ACC, and 1.96\% in F-score compared to RGs-UMC, highlighting the effectiveness of our method in learning confident cluster structures. In contrast, RGs-UMC fails to account for the confidence in cluster structures across views, resulting in inferior results.

\begin{table*}
\centering
\vspace{-1mm}
\renewcommand{\arraystretch}{1.1}
\setlength{\tabcolsep}{4.5pt}
\caption{Comparison of related multi-view clustering methods for UMC on \textbf{two views} datasets.}\label{table4_comparison}
\begin{tabular}{l|ccc|ccc|ccc|ccc|ccc}
\hline\hline
\multirow{2}{*}{\textbf{Methods}}& \multicolumn{3}{c}{\emph{Digit}} \vline &\multicolumn{3}{c}{\emph{Scene-15}} \vline &\multicolumn{3}{c}{\emph{Caltech101-20}} \vline
&\multicolumn{3}{c}{\emph{Flower17}}\vline
&\multicolumn{3}{c}{\emph{{Reuters}}}\\ \cline{2-16}
&\emph{NMI} &\emph{ACC} &\emph{F1} &\emph{NMI} &\emph{ACC} &\emph{F1} &\emph{NMI} &\emph{ACC} &\emph{F1} &\emph{NMI} &\emph{ACC} &\emph{F1}&\emph{NMI} &\emph{ACC} &\emph{F1}\\\hline
\textbf{OPMC} {(ICCV2021)
}&44.12	&51.75	&32.02	&30.43	&27.29	&19.76	&27.59	&28.12	&24.71	&26.19	&24.56	&14.91 &20.17	&40.32	&{36.73}\\ 
\textbf{OP-LFMVC} {(ICML2021)}&39.04	&46.40	&35.10	&15.60	&20.45	&13.28	&21.61	&22.09	&17.46	&20.81	&23.31	&13.09 &- &- &-\\ 
\textbf{DUA-Nets} {(AAAI2021)} &42.08 &38.07 &33.98 &30.12	&27.80	&21.23 &32.40	&29.26	&25.42 & 28.34 &24.44	&18.61 &4.28 &26.23	&24.17\\
\textbf{RMVC} {(AAAI2018)}&46.75  &44.05 &36.56 &36.34 &30.52 &21.92  &34.37	&24.43   &22.11 &33.70	&28.24	&18.83 &- &- &-\\
\textbf{DSMVC} {(CVPR2022)}&49.84 &41.90	&8.39 &27.00	&25.22	&6.03 &32.14& 22.34&5.29 &26.69 &23.46 &4.14 &10.86 &	30.39 &14.40\\
{\textbf{AWMVC} {(AAAI2023)}} &41.45	&44.05	&31.76  &30.39 &28.63	&20.17 &33.11	&33.74	&29.24 &27.76	&26.19	&16.31 &24.43	&36.70	&35.01\\
{\textbf{UOMvSC} {(TKDE2023)}} &52.56	&41.95	&37.09 &34.81	&30.01	&21.99 &35.18	&26.45	&22.72 &29.28	&26.18	&16.43 &- &- &- \\
{WMVEC-FP} {(KBS2024)}&37.55	&41.75	&22.12 &27.31	&30.68	&22.97 &21.37	&21.56	&12.96 &13.35	&13.01	&10.07  &11.98	&31.09	&30.92 \\
{CVCL} {(ICCV2023)} &69.07	&73.70	&73.32 &38.41	&33.06	&29.65   &47.75	&39.07	&26.79    &31.84	&32.16	&30.63 &- &- &-\\ \hline
\textbf{DAIMC} {(IJCAI2018)}&39.37	&38.50	&30.88	&19.46&	22.79	&16.20	&20.16	&24.81	&22.88	&21.87	&21.03	&15.12 &- &- &- \\ 
\textbf{UEAF} {(AAAI2019)} &30.90	&26.77	&19.32	&22.16	&20.14	&13.19	&25.44	&22.19 	&32.54	&12.72	&9.25	&6.37 &- &- &-\\ 
\textbf{IMSC-AGL} {(TCYB2020)}&48.99 &45.40 &37.09&27.61	&27.98	&18.33 &30.04	&24.06	&20.38 &26.99 &24.71 &15.31  &- &- &-\\
\textbf{OMVC} {(BigData2017)} &31.69	&34.87	&20.64 & 0.68& 9.16&12.94 &26.11&26.78	&22.62 &26.08	&24.52	&14.47 &-&-&-
\\ 
\textbf{OPIMC} {(AAAI2019)}&41.49 &44.60 &37.01 &20.32&23.52&15.30 &22.38&24.48&21.97 &19.90&21.84&12.22 &8.99	&27.53	&24.36\\ 
\textbf{MvCLN} {(CVPR2021)}&41.96	&45.10	&52.41 & 21.50	&21.43	&22.47 &31.21	&28.04	&17.64 &19.57	&17.65	&18.43 &20.18	&18.97	&19.72\\
\textbf{Completer} {(CVPR2021)}&41.35 &39.66	&40.24	&23.24	&21.86	&19.59	&34.29	&26.37	&22.61	&22.65	&20.73	&19.83 &1.28	&23.84	&19.67\\
{\textbf{FCMVC} {(TIP2024)}} &37.31	&45.65	&32.55 &17.21	&23.63	&13.40 &24.55	&21.04	&17.47 &14.91	&18.16	&9.96 &11.75	&39.12	&27.73\\
\textbf{T-UMC} {(TCYB2022)}&46.15	&49.85	&49.26 &40.75	&39.20	&30.10 &38.84	&35.88	&23.54 &22.44	&22.65 &16.51 &-&-&-\\
{\textbf{GIGA} } {(PR2024)} &52.17	&40.45	&41.62 &13.25	&14.72	&12.84 &47.73	&33.49	&26.30 &24.67	&20.96	&20.53 &-&-&-\\
\textbf{{ICMVC}} {(AAAI2024)} &25.54&30.20&30.39 &12.62&18.02&17.67
    &33.86&21.46&17.31    &35.49&27.72&26.98  &13.84&35.60&26.00  \\\hline
\textbf{{UPMGC-SM}} {(TNNLS2024)}&39.41&38.46&28.99 &25.42&24.38&14.72 &23.38&20.88&19.25 &23.60&23.72&14.09 &- &- &-\\

\textbf{IUMC-CA} {(TNNLS2023)}&46.20	&27.75	&29.82	&{40.50}	&30.08	&27.39	&35.91	&26.40	&19.24	&\bf{43.31}	&{31.84}	&25.64 &- &- &-\\ 
\textbf{IUMC-CY} {(TNNLS2023)} &49.17	&45.55	&37.72	&39.67	&18.68	&23.87	&32.22	&28.08	&{35.06}	&39.52	&18.82	&21.01 &11.34	&27.50	&27.36\\
    
\textbf{scl-UMC} {(TNNLS2023)}&{57.00}&{60.85}&{61.35}	&{45.26}	&{36.59}	&{35.35} &{38.60}	&{41.52}	&{36.84}&{39.98}	&{32.21}	&{35.06} & {39.50}	&{49.86}	&{41.12}\\

{\textbf{MGCCFF}} {(AAAI2025)} &{65.66}	&{68.79}	&{60.41}	&{36.79} &{34.59}	&{25.97}	&{60.75}	&{43.66}	&{48.28} &{29.21}	&{27.53}	&{19.25} & {-} & {-} & {-}\\

\textbf{{SURE}} {(TPAMI2022)}&64.93  &52.05  &54.18 &48.03 &36.68  &36.50   &64.82  &43.04  &32.34 &25.77 &24.34   &23.79 &53.59  &45.88   &42.42\\ 
\textbf{RG-UMC} {(TNNLS2024)}&{92.38}&{96.40}&{96.42}	&{46.79}	&{49.53}	&{47.40} &\bf{78.64}	&\bf{73.89}	&\bf{55.46}&{42.01}	&\underline{43.75}	&\underline{42.80}	&{53.56}	&{60.53}	&{39.13}\\
\textbf{RGs-UMC} {(TNNLS2024)} &\underline{93.93}&\underline{97.20}& \underline{97.21}    &\bf{48.57}&\underline{52.19}&\underline{50.70}    &\underline{76.37}&{71.17}&{54.50}  &{41.35}&{41.62}&{42.30}   &\underline{55.82}&\bf{71.96}&\underline{56.18}\\
\textbf{MRG-UMC} &\bf{94.11}	&\bf{97.30}	&\bf{97.31}  &\underline{47.90}	&\bf{53.86}	&\bf{51.22}  &76.19&\underline{71.52}&\underline{54.58}  &\underline{42.66}	&\textbf{43.75}	&\bf{42.83} &\bf{56.62} &\underline{63.43} & \bf{56.40} \\
\hline\hline
\end{tabular}
\vspace{-0.5cm}
\end{table*}

\textbf{Clustering performance comparison of methods on two-view datasets.} 
Following \cite{lin2021completer}, we evaluate our method on two-view datasets. For the five datasets, \emph{Digit} and \emph{Flower17} use their first two views as two-view datasets, while the remaining datasets follow the settings from \cite{lin2021completer, scl-UMC}.
TABLE \ref{table4_comparison} summarizes the results against 27 methods, leading to the following observations:
(i) Compared with MC and IMC methods, MRG-UMC consistently outperforms all of them, demonstrating superior effectiveness in UMC.
(ii) MRG-UMC also surpasses the other UMC methods, achieving average improvements of 15.56\% in NMI, 18.23\% in ACC, and 18.69\% in F-score. 
These improvements are attributed to the multi-level clustering mechanism, which reduces clustering errors, as well as the modules designed to enhance the learning of consistent and confident cluster structures.

\begin{table*}
\centering
\renewcommand{\arraystretch}{1.1}
\setlength{\tabcolsep}{6.5pt}
\caption{{Compare single-level and multi-level clustering of MRG-UMC on four datasets with multiple views.}}\label{single-multi-level}
\vspace{-0.2cm}
\begin{tabular}{c|ccc|ccc|ccc|ccc}
\hline\hline
\multirow{2}{*}{\textbf{{Strategy}}} &\multicolumn{3}{c}{\emph{{Digit (6views)}}} \vline &\multicolumn{3}{c}{\emph{{Scene-15 (3views)}}} \vline &\multicolumn{3}{c}{\emph{{Caltech101-20 (6views)}}} \vline &\multicolumn{3}{c}{\emph{{Flower17 (7views)}}} \\ \cline{2-13}%
& \emph{NMI} &\emph{ACC} & \emph{F1} &\emph{NMI} & \emph{ACC} & \emph{F1} & \emph{NMI} & \emph{ACC} & \emph{F1} & \emph{NMI} & \emph{ACC} & \emph{F1}\\ \hline
{single-level clustering} &86.60	&92.93	&92.86	&44.34	&49.07	&45.86	&71.05	&75.32	&38.62	&59.72	&52.52	&51.95\\
{multi-level clustering} &88.36	&93.99	&93.92	&49.42	&53.88	&50.47	&72.82	&76.48	&42.01	&61.76	&54.85	&53.33\\		
\hline\hline
\end{tabular}
\vspace{-0.5cm}
\end{table*}

\subsection{{{Validation of Two Propositions}}}
To evaluate the impact of multi-level clustering described in Proposition 1, we conducted two experimental analyses.

{\textbf{(1)The performance between single-level and multi-level clustering.}}
{To validate Proposition 1, we compare single-level clustering and multi-level clustering on four datasets. The results from Table \ref{single-multi-level} show that multi-level clustering performance outperforms its single-level counterpart, confirming the effectiveness of proposition 1. Specifically, on four datasets, multi-level clustering achieves an average improvement of 2.66\%, 2.34\%, and 2.61\% over single-level clustering in terms of NMI, ACC, and F1, respectively.}

\begin{figure}[!t]
\centering
\includegraphics[width=0.95\columnwidth]{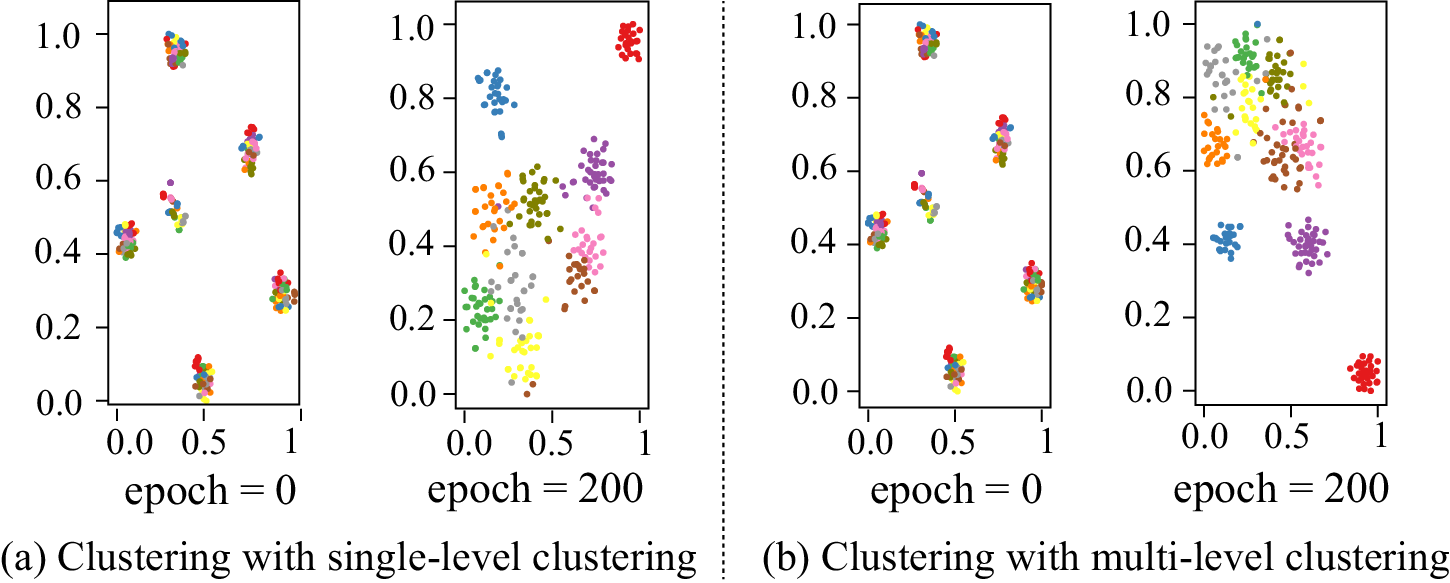}
\vspace{-0.3cm}
\caption{{Evolution of the latent representation on the \emph{Digit} dataset (6 views and 10 clusters) for Proposition 1. Each color denotes a cluster.}}
\label{digit20250414-clustering-propostion1}
\vspace{-0.5cm}
\end{figure} 

{\textbf{(2) Visualization the clustering between single-level and multi-level clustering.}}
{To further demonstrate the effectiveness of multi-level clustering, we visualize the clustering evolution on the latent representations of the \emph{Digit} dataset with six views, as shown in Fig. \ref{digit20250414-clustering-propostion1}. By comparing Fig. \ref{digit20250414-clustering-propostion1}(a) and Fig. \ref{digit20250414-clustering-propostion1}(b), we observe that multi-level clustering improves sample separability and compactness, with fewer samples incorrectly merged near decision boundaries, which is consistent with the analysis of Proposition 1 in Fig. \ref{Venn_diagram}.}

\begin{table*}
\centering
\renewcommand{\arraystretch}{1.1}
\setlength{\tabcolsep}{6.5pt}
\caption{{Performance of MRG-UMC with and without reliable view guidance on four datasets.}}\label{with-without-reliable-view}
\vspace{-0.2cm}
\begin{tabular}{c|ccc|ccc|ccc|ccc}
\hline\hline
\multirow{2}{*}{\textbf{{Strategy}}} &\multicolumn{3}{c}{\emph{{Digit (6views)}}} \vline &\multicolumn{3}{c}{\emph{{Scene-15 (3views)}}} \vline &\multicolumn{3}{c}{\emph{{Caltech101-20 (6views)}}} \vline &\multicolumn{3}{c}{\emph{{Flower17 (7views)}}} \\ \cline{2-13}%
& \emph{NMI} &\emph{ACC} & \emph{F1} &\emph{NMI} & \emph{ACC} & \emph{F1} & \emph{NMI} & \emph{ACC} & \emph{F1} & \emph{NMI} & \emph{ACC} & \emph{F1}\\ \hline
{w/o reliable view guidance} &17.44	&22.58	&23.39 &19.05	&18.96	&17.46 &44.34	&49.07	&45.86 &15.89	&14.82	&13.30\\
{with reliable view guidance} &88.36	&93.99	&93.92	&49.42	&53.88	&50.47	&72.82	&76.48	&42.01	&61.76	&54.85	&53.33\\
\hline\hline
\end{tabular}
\vspace{-0.5cm}
\end{table*}

{To validate the effectiveness of reliable view guidance as described in Proposition 2, we conducted three experimental analyses.}

{\textbf{(1) Performance of reliable view guidance.}}
{We evaluate the clustering performance with and without reliable view guidance, as shown in Table \ref{with-without-reliable-view}. From the results in Table \ref{with-without-reliable-view}, we can clearly observe the significant impact of reliable view guidance.}

{\textbf{(2) Individual view performance under reliable view guidance.}}
{Furthermore, we examine the impact of reliable view guidance on each individual view, with detailed results presented in Table \ref{sing-level}. The results demonstrate that reliable view guidance effectively enhances clustering performance at the view level.}
\begin{table}
\centering
\renewcommand{\arraystretch}{1.1}
\setlength{\tabcolsep}{6.5pt}
\caption{{The clustering performance of each view with and without reliable view guidance}}\label{sing-level}
\vspace{-0.2cm}
\begin{tabular}{c|ccc|ccc}
\hline\hline
\multirow{2}{*}{{{View}}} &\multicolumn{3}{c}{\emph{{Digit (6views)}}} \vline  &\multicolumn{3}{c}{\emph{{Caltech101-20 (6views)}}}  \\ \cline{2-7}%
& \emph{NMI} &\emph{ACC} & \emph{F1} &\emph{NMI} & \emph{ACC} & \emph{F1} \\ \hline
\multicolumn{7}{c}{{{W/o reliable view guidance}}}\\\hline
{V1} &42.95	&47.58	&46.70    &32.21	&31.62	&19.68	\\
{V2} &70.42	&66.97	&64.07   &39.19	&38.30	&21.04	\\		
{V3} &47.23	&57.88	&55.61     &33.46	&40.87	&19.74		\\
{V4} &65.29	&59.70	&58.20   &57.82	&55.01	&23.62	\\		
{V5} &53.82	&58.18	&56.80   &46.40	&36.25	&20.02	 \\
{V6} &66.76	&63.03	&59.73    &44.24	&39.33	&18.62	\\\hline

\multicolumn{7}{c}{{{With reliable view guidance}}}\\\hline
{V1} &88.07	&88.79	&88.29   &66.95	&58.35	&39.76\\
{V2} &97.63 &98.79	&98.79   &61.76	&46.79	&34.33\\		
{V3} &93.50	&96.36	&96.23   &65.53	&48.59	&34.11\\
{V4} &97.05 &98.48	&98.48   &66.18	&47.04	&32.32\\		
{V5} &89.84	&84.55	&82.35   &70.73	&64.78	&40.58\\
{V6} &75.31	&76.67	&72.03   &70.83	&60.93	&43.25\\		
\hline\hline
\end{tabular}
\vspace{-0.3cm}
\end{table}

\begin{figure}[!t]
\centering
\includegraphics[width=0.95\columnwidth]{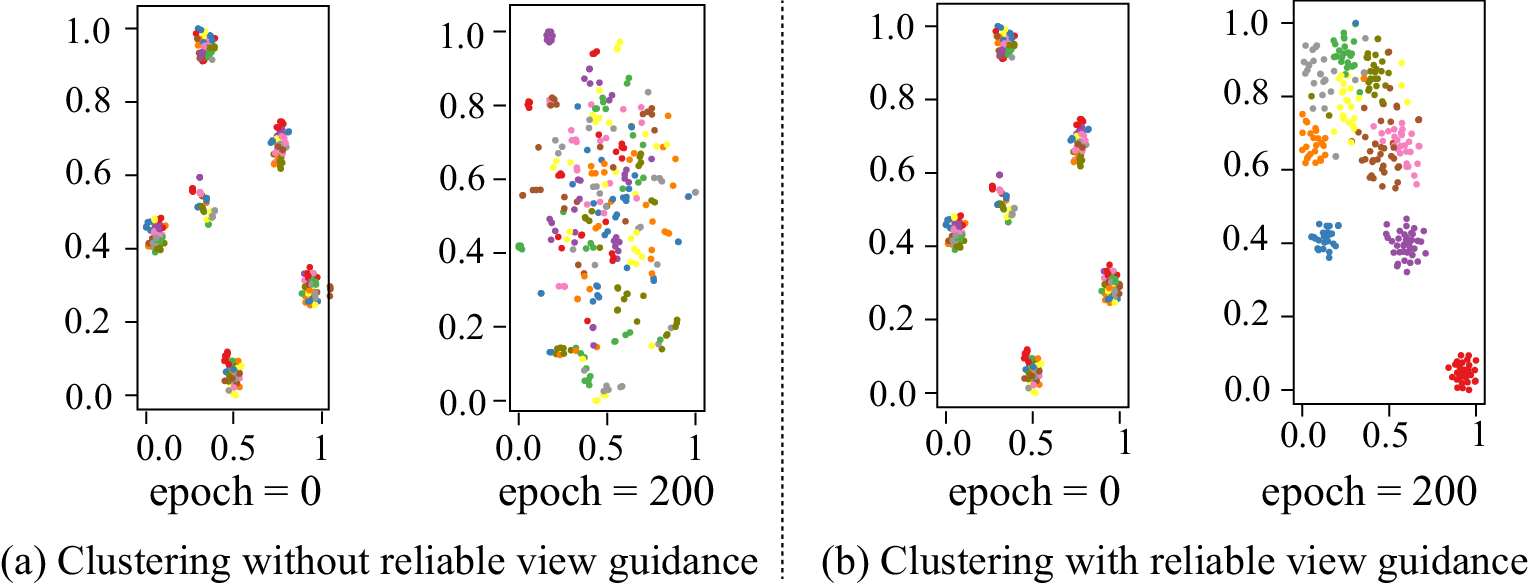}
\vspace{-0.3cm}
\caption{{Evolution of the latent representation on the \emph{Digit} dataset (6 views and 10 clusters) for Proposition 2. Each color denotes a cluster.}}
\label{digit20250414-clustering-propostion2}
\vspace{-0.5cm}
\end{figure}

{\textbf{(3) Visualization of clustering under reliable view guidance.}}
{To further illustrate the effectiveness of reliable view (\ie cross-view) guidance, we visualize the clustering evolution in Fig. \ref{digit20250414-clustering-propostion2}. Comparing panels (a) and (b), we observe that in the absence of cross-view guidance, the clustering structure becomes noticeably less coherent. Furthermore, comparing Fig. \ref{digit20250414-clustering-propostion1} with Fig. \ref{digit20250414-clustering-propostion2}, the results suggest that cross-view guidance contributes more significantly than multi-level clustering to the formation of well-structured clusters.}

\subsection{Ablation Study and Visualization}

\begin{table*}
\centering
\renewcommand{\arraystretch}{1.1}
\setlength{\tabcolsep}{6.5pt}
\caption{Ablation analysis of MRG-UMC conducted on four datasets.}\label{ablationstudy-Digit}
\vspace{-0.2cm}
\begin{tabular}{c|cccc|ccc|ccc|ccc|ccc}
\hline\hline
\multirow{2}{*}{\textbf{Line}}&\multirow{2}{*}{\emph{Orth}} &\multirow{2}{*}{\emph{In}} &\multirow{2}{*}{\emph{Co}} &\multirow{2}{*}{\emph{Cr}} &\multicolumn{3}{c}{\emph{Digit (6views)}} \vline &\multicolumn{3}{c}{\emph{Scene-15 (3views)}} \vline &\multicolumn{3}{c}{\emph{Caltech101-20 (6views)}} \vline &\multicolumn{3}{c}{\emph{Flower17 (7views)}} \\ \cline{6-17}%
& & & & & \emph{NMI} &\emph{ACC} & \emph{F1} &\emph{NMI} & \emph{ACC} & \emph{F1} & \emph{NMI} & \emph{ACC} & \emph{F1} & \emph{NMI} & \emph{ACC} & \emph{F1}\\ \hline
1&\quad&  \quad & \quad  & \quad     & 15.12	&20.86	&21.19 &20.81	&20.28	&18.44 & 19.40	&23.69	&12.57 & 15.45& 13.98& 14.20 \\
2&\checkmark&  \quad & \quad  & \quad & 15.23 &21.67	&22.36 &21.40	&21.02	&19.27 & 20.48	&27.59	&13.33 & 13.48	&13.14	&13.47  \\
3&\quad &\checkmark&  & \quad         &15.07	&22.07	&23.04 &21.01	&20.37	&18.64 & 18.10	&41.39	&12.21 &11.75	&11.61	&11.76 \\
4&\quad& & \checkmark & \quad         &14.88	&21.52	&21.92 & 21.15	&21.00	&20.44 &23.35	&21.81	&13.86 &14.73	&14.06	&13.10 \\
5&\quad&  \quad & \quad & \checkmark  &83.04	&90.45	&90.35 & 41.44	&47.41	&42.87 &64.14	&64.82	&34.27 &38.95	&36.90	&36.00 \\
6&\checkmark&  &  & \checkmark        &85.27	&92.07	&91.99 & 42.06	&47.84	&44.96 &70.02	&67.35	&36.75 &56.04	&48.66	&45.67 \\
7& &\checkmark  &  &\checkmark        &85.30	&92.22	&92.12  & 42.02	&46.97	&45.73 &67.68	&65.21	&36.51 &43.12	&38.81	&34.42 \\
8& & &\checkmark & \checkmark         &85.44	&92.22	&92.14 & 41.48	&46.83	&43.03 &70.42	&74.29	&38.64 &47.52	&43.54	&40.45 \\
9&\checkmark& \checkmark &   & \checkmark  &85.57	&92.47	&92.41 & 43.01	&48.11	&44.11 &70.74	&75.49	&38.90 &55.94	&50.50	&48.79 \\
10&\checkmark&   & \checkmark & \checkmark &85.47	&92.22	&92.13 & 42.41	&50.93	&48.13 &69.26	&70.69	&38.21 &54.63	&49.81	&47.00 \\
11& &\checkmark  &\checkmark & \checkmark  &85.84	&92.22	&92.10 & 41.09	&49.27	&45.70 &70.78	&73.35	&39.98 &47.57	&45.30	&43.72 \\
12&\checkmark &\checkmark &\checkmark & \checkmark &\textbf{88.36}	&\textbf{93.99}&\textbf{93.92} & \textbf{46.09}	& \textbf{53.08}	&\textbf{50.47} &\textbf{72.82}	&\textbf{76.48}&\textbf{42.01} &\textbf{61.76}	&\textbf{54.85}&\textbf{53.33} \\		
\hline\hline
\end{tabular}
\vspace{-0.3cm}
\end{table*}

\textbf{Ablation study of MRG-UMC.}
To further analyze the modules in MRG-UMC, we conducted ablation studies on four datasets (\emph{Digit}, \emph{Scene-15}, \emph{Caltech101-20} and \emph{Flower17}). 
Specifically, we evaluated different combinations of the orthogonal constraint ($Orth$), the inner-view multi-level clustering ($In$), the synthesized-view alignment module ($Sy$), and the cross-view guidance module ($Cr$). The results of the ablation studies are presented in TABLE \ref{ablationstudy-Digit}.

As observed from TABLE \ref{ablationstudy-Digit}, the combination of all modules performs the best (Line 12), which is our method MRG-UMC. In detail, when comparing Lines 2-5 with Line 1, it is evident that the $Cr$ module has the most significant impact on clustering, validating the importance of reliable view guidance. 
Subsequently, the following ablations are based on the $Cr$ module. 
Taking the \emph{Digit} dataset as an example, when comparing Lines 6-8 with Line 5, it is evident that the $Orth$ module and $Sy$ module are more effective than the $In$ module in learning the cluster structure. 
The $Orth$ module aids in learning discriminative features, while the $Sy$ module facilitates to alignment of different views. 
Besides, the $In$ module helps to progressively merge confident cluster structures through multi-level clustering.
Moreover, comparing Lines 9-11 with Lines 6-8 shows that these three modules complement each other and improve performance. The same conclusion can be drawn from other datasets in TABLE \ref{ablationstudy-Digit}.

\begin{figure*}[!t]
\centering
\includegraphics[width=1.5\columnwidth]{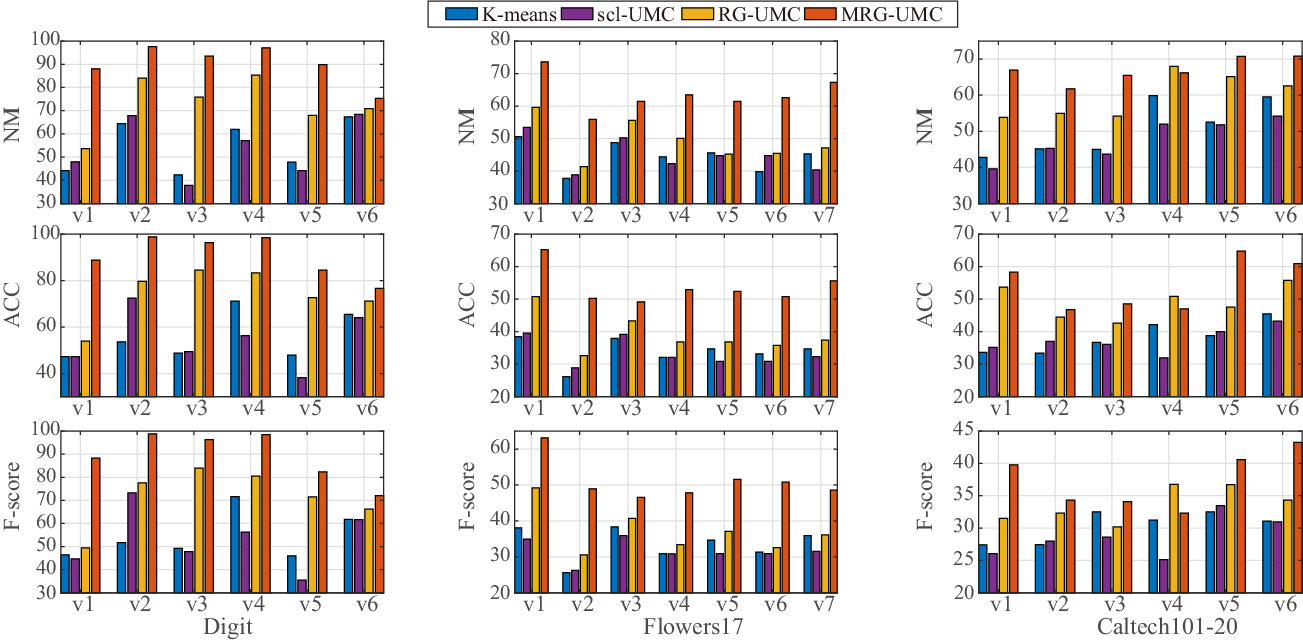}
\vspace{-0.3cm}
\caption{{The single-view performance of three comparison methods across three datasets.}}
\label{super-class-singleview}
\vspace{-0.5cm}
\end{figure*}

\begin{figure*}[!t]
\centering
\includegraphics[width=1.95\columnwidth]{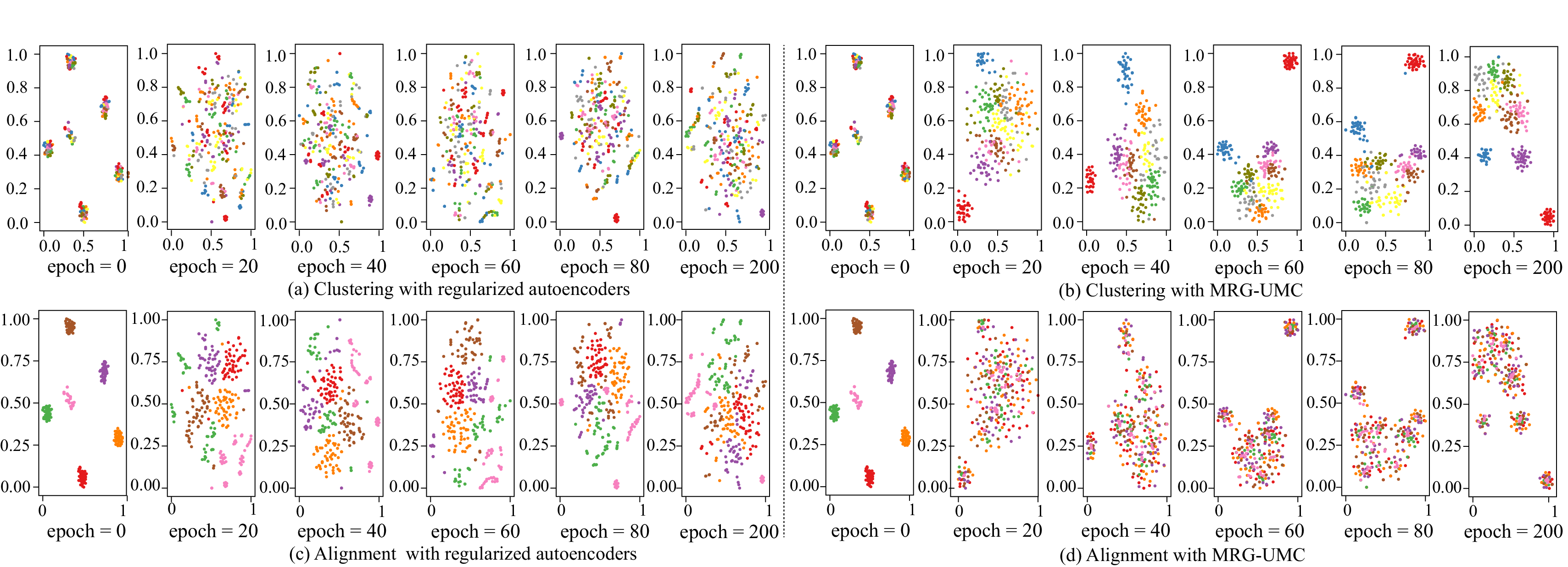}
\vspace{-0.3cm}
\caption{{Comparison of visualizations between MRG-UMC and the MRG-UMC model with only regularized autoencoders. The evolution of latent representations on the \emph{Digit} dataset with ten clusters and six views during the training process. In the first and second rows, colors represent clusters and views, respectively.}}
\label{digit-tsne}
\vspace{-0.5cm}
\end{figure*}

\textbf{Single-view performance.} 
MRG-UMC utilizes a consistent cluster structure to connect different views. To validate the effectiveness of MRG-UMC on each view, we evaluate its performance across three datasets and compare it with three other methods, as shown in Fig. \ref{super-class-singleview}.
i) Comparing the $K$-means clustering method with all the other methods validates the effectiveness of exploring a consistent cluster structure.
ii) Comparing scl-UMC and RG-UMC with MRG-UMC, we observed significant improvements in clustering performance for each view. {This shows that a more confident cluster structure helps to improve clustering performance.}

\textbf{Visualization.} To show the effectiveness of learning consistent cluster structures and achieving alignment, we use t-SNE to track the evolution every 20 epochs during training on the \emph{Digit} dataset with multiple views. For clarity, we randomly sample 500 samples and visualize the evolution in Fig. \ref{digit-tsne}. In Fig. \ref{digit-tsne} (a-b), colors represent cluster assignments predicted by $K$-means, while in Fig. \ref{digit-tsne} (c-d), colors indicate different views. Specifically, to highlight the effectiveness of our method, we compare the visualizations of MRG-UMC (Fig. \ref{digit-tsne} (b) and (d)) with those of the MRG-UMC model with only regularized autoencoders (\ie Eq. (\ref{eq1})) (Fig. \ref{digit-tsne} (a) and (c)).

From Fig. \ref{digit-tsne}, we observe:
i) Comparing Fig. \ref{digit-tsne} (a) and (b), the three modules of MRG-UMC aid in forming cluster structures.
Similarly, comparing Fig. \ref{digit-tsne} (c) and (d), MRG-UMC efficiently and quickly aligns different views.
ii) From Fig. \ref{digit-tsne} (b), we see that initially, features are mixed, with most samples grouped into a few clusters. As training progresses, cluster assignments improve, and features become more distinct. This demonstrates the effectiveness of MRG-UMC in learning clear cluster structures.
iii) Fig. \ref{digit-tsne} (d) shows that our method successfully aligns samples from different views. Initially, the samples are separated by view, but as training goes on, they quickly come into alignment.

\begin{table}
\centering
\renewcommand{\arraystretch}{1.1}
\setlength{\tabcolsep}{2.5pt}
\caption{{FLOPs and total parameter of MRG-UMC on four datasets with multiple views}}\label{FLOPs-total-parameter}
\vspace{-0.2cm}
\begin{tabular}{c|cccc}
\hline\hline
\multirow{2}{*}{\textbf{{Datasets}}} &{\emph{Digit}} &{\emph{Scene15 }} &{\emph{Caltech101-20}}  &{\emph{Flower17}} \\
&{(6views)} &{(3views)} &{(6views)} &{(7views)} \\\hline
{Total params (MB)} &4.63  &4.52  &4.57   &7.26   \\
{GFLOPs } &1.18  &1.15    &1.17    &1.36\\\hline\hline		
\end{tabular}
\vspace{-0.3cm}
\end{table}

{\textbf{Complexity analysis.}}
{We analyze the computational complexity of MRG-UMC by computing the number of parameters and floating-point operations (FLOPs) \cite{shi2023deep, Chen2025AStatic}  across four datasets in Table \ref{FLOPs-total-parameter}. Across the four datasets, MRG-UMC maintains a total parameter size below 10 MB and requires less than 1.5 GFLOPs, highlighting its efficiency and suitability for scalable or real-time clustering applications.}

\begin{table}
\centering
\renewcommand{\arraystretch}{1.1}
\setlength{\tabcolsep}{1.3pt}
\caption{{External indicators evaluation on two datasets}}\label{External-indicators-evaluation}
\vspace{-0.2cm}
\begin{tabular}{c|cc|ccc|cc|ccc}
\hline\hline
\multirow{2}{*}{\textbf{{Methods}}} &\multicolumn{5}{c}{{\emph{Digit (6views)}}}  \vline &\multicolumn{5}{c}{{\emph{Caltech101-20 (6views)}}}\\ \cline{2-11}
& NMI & ACC & sil $\uparrow$ &DB $\downarrow$ & CHI $\uparrow$ & NMI & ACC & sil $\uparrow$ &DB $\downarrow$ & CHI $\uparrow$\\\hline
{\textbf{WMVEC-FP}}	&10.75	&18.25  &-0.07	&\underline{2.93}	&34.94	&17.19	&19.04	 &\bf{0.62}	&\underline{2.42}	&\underline{85.95}\\
{\textbf{FCMVC}} &12.39	&22.55	 &-0.45	&17.89	&33.33	&9.73	&13.29	&-0.26	&13.03	&5.83\\
{\textbf{GIGA}} &8.84	&17.50	&\underline{0.10}	&5.85	&11.35	&21.16	&16.26	&-0.01	&5.26	&10.02\\
{\textbf{UPMGC-SM}} &31.43&35.36 &-0.30	&20.66	&34.52	&16.84&16.49 &-0.69	&16.60	&15.62\\
{\textbf{RG-UMC}} &{77.37} &{84.85} &0.03	&4.79	&42.09	&{65.81}&{67.05} &0.07	&5.16	&26.04\\
{\textbf{RGs-UMC}} &\underline{86.19}&\underline{92.83} &0.04	&4.19	&\underline{43.52}	&\underline{70.88}&\underline{75.28} &0.09	&5.03	&30.74\\
{\textbf{MRG-UMC}} &\textbf{88.36}	&\textbf{93.99}  &\bf{0.15}	&\bf{2.53}	&\bf{192.48}	&\textbf{72.82}	&\textbf{76.48}	&\underline{0.19}	&\bf{2.41}	&\bf{86.31}
\\\hline\hline
\end{tabular}
\vspace{-0.5cm}
\end{table}

{\textbf{Cluster evaluation validity index.}}
{To evaluate the effectiveness of MRG-UMC in terms of clustering quality, we compare it with six state-of-the-art methods published in 2024 using three widely adopted internal metrics: Silhouette Coefficient (sil) \cite{lin2022tensor}, Davies-Bouldin Index (DB), and Calinski-Harabasz Index (CHI) \cite{zhong2020analysis}. 
The Silhouette Coefficient ranges from $[-1, 1]$, with higher values indicating a better clustering structure.
The DB Index ranges from $0$ to $+\infty$, where lower values are preferred. The CHI Index also ranges from $0$ to $+\infty$, with higher values indicating better cluster compactness and separation.}
{Experiments are conducted on two representative unpaired multi-view datasets, \emph{Digit} and \emph{Caltech101-20}, with results summarized in Table \ref{External-indicators-evaluation}. As shown, MRG-UMC consistently outperforms other methods across most metrics, demonstrating more semantically meaningful clustering with improved intra-cluster compactness and inter-cluster separability. In contrast, some methods (\eg WMVEC-FP) yield high internal scores (sil, DB, and CHI) and exhibit clear cluster structures, yet perform poorly on external metrics (NMI and ACC), suggesting a misalignment between the learned clusters and the ground truth labels.}

\section{Conclusion and Future Work}
Mining consistency among views is challenging when the cluster structures with low confidence. To address this issue, we propose a novel method called Multi-level Reliable Guidance for UMC (MRG-UMC) to enhance the confidence of cluster structures. Specifically, MRG-UMC consists of three modules: inner-view multi-level clustering, synthesized-view alignment, and cross-view guidance. Ultimately, our method demonstrates outstanding performance in UMC, as evidenced by both theoretical analysis and extensive test results.

In future work, we plan to enhance the robustness of multi-view models. For the ever-changing reality, ensuring the reliability, security, and cleanliness of data becomes increasingly challenging. Furthermore, while many methods aim to enhance overall performance, they often neglect weaker performance on individual views. Consequently, we will further investigate algorithms within stable and secure frameworks.

\bibliographystyle{IEEEtran}
\bibliography{IEEEabrv,egbib2}

\end{document}